\documentclass{article}
\usepackage{iclr2027_conference,times}

\usepackage{amsmath,amsfonts,bm}

\def\eqref#1{equation~\ref{#1}}

\def\1{\bm{1}}

\DeclareMathAlphabet{\mathsfit}{\encodingdefault}{\sfdefault}{m}{sl}
\SetMathAlphabet{\mathsfit}{bold}{\encodingdefault}{\sfdefault}{bx}{n}

\usepackage{amssymb}
\usepackage{graphicx}
\usepackage{booktabs}
\usepackage{colortbl}
\usepackage{multirow}
\usepackage{tabularx}
\usepackage{array}
\usepackage{algorithm}
\usepackage{algpseudocode}
\usepackage{hyperref}
\usepackage{url}
\definecolor{odaRow}{RGB}{237,238,255}
\definecolor{odaRate}{RGB}{58,68,138}
\newcommand{\odaavgrate}[1]{{\scriptsize\textcolor{odaRate}{\, (#1\%)}}}

\hypersetup{
  pdftitle={On-Demand Attention: Language Models Know When to Recall},
  pdfauthor={Haibo Feng, Ruiqi Liang, Dongyang Jin, Hanyang Peng, Shiqi Yu}
}
\title{On-Demand Attention:\\ Language Models Know When to Recall}
\author{
Haibo Feng\textsuperscript{1,3}, Ruiqi Liang\textsuperscript{2}, Dongyang Jin\textsuperscript{1}, Hanyang Peng\textsuperscript{3}, Shiqi Yu\textsuperscript{1}\\
\normalfont\textsuperscript{1}Southern University of Science and Technology\\
\normalfont\textsuperscript{2}Peking University\\
\normalfont\textsuperscript{3}Peng Cheng Laboratory
}
\iclrfinalcopy
\hypersetup{hidelinks}
\begin{document}
\maketitle
\pagestyle{plain}
\thispagestyle{plain}

\begin{abstract}
Reasoning and agentic workloads increasingly demand efficient long-context inference.
Yet full-attention decoding reads the growing history at every step, although the benefit of global access varies across prediction positions.
We find that, before global attention is computed for the current step, the decoding states available after local computation in frozen pretrained models already contain information predictive of its benefit over local attention.
Building on this finding, we introduce On-Demand Attention (ODA), a local-first decoding method: after local computation, a lightweight recall head decides whether to recompute the current step with global attention.
ODA trains only the recall head with modest data and compute budgets, leaving pretrained weights unchanged and retaining the complete historical KV cache so that information skipped at one step remains available for later access.
Experiments across model scales and families, including hybrid attention backbones, show that ODA recovers most of the performance lost under local attention while substantially reducing the frequency of global attention.
Controlled long-context measurements in vLLM further show that GPU-side conditional execution translates fewer global reads into practical decoding speedups over full attention.
These findings show that pretrained decoding states can support both token prediction and decisions about accessing distant information, allowing models to allocate global computation as needed during decoding.
Code will be available upon acceptance.
\end{abstract}

\section{Introduction}
\label{sec:introduction}

Reasoning and agentic tasks often require large language models to generate extended outputs while processing a growing history, making efficient long-context decoding increasingly important~\citep{anthropic2026cadences,longspec2026}.
Full-attention decoding accesses the complete history at every step, so history reads and attention computation become more costly as the context grows~\citep{quest2024}.
However, the additional benefit of accessing distant information over using only the local context can vary across generation steps~\citep{shortcontextdominance2026,aha2025}.
Invoking global attention at every step does not account for this variation and incurs the cost of reading the complete history even when the additional benefit is limited.
This raises a key question: before computing global attention for the current step, can we estimate the additional predictive benefit it would provide over local computation and use this estimate to decide whether to access the complete history?

We find that the decoding states available after local computation in pretrained models already contain signals predictive of the benefit of global attention over local attention (Figure~\ref{fig:budget-timing}(a)).
This gives local computation a dual role: its output supports the current prediction and, together with other available states, helps determine whether additional global computation is worthwhile.
One class of dynamic attention methods jointly trains the language model and its access policy to learn local--global allocation~\citep{aha2025,l2a2026,switch2026,logo2026}; we instead keep pretrained weights unchanged and learn to read out this benefit signal from the model's existing states to guide global access.

\begin{figure}[t]
    \centering
    \includegraphics[width=\linewidth]{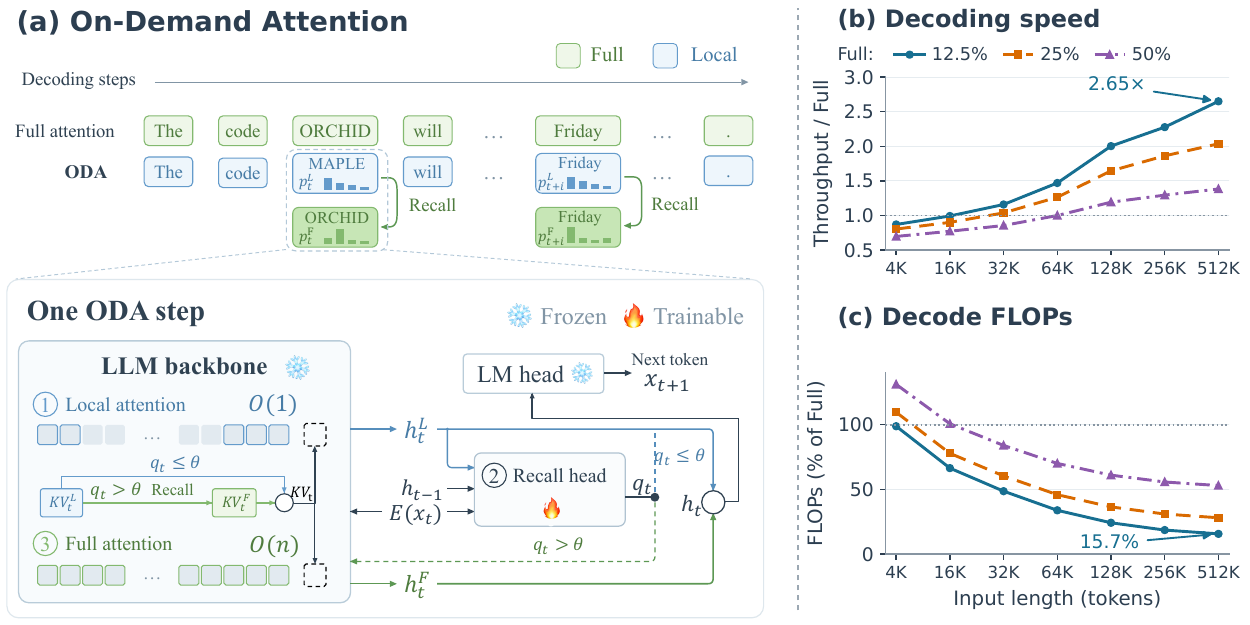}
    \caption{\textbf{ODA decoding and execution efficiency.}
    \textbf{(a)} After Local computation, the recall head accepts its result
    or triggers Full recomputation from the same pre-step history. Recall
    updates the current next-token distribution from $p_t^L$ to $p_t^F$
    before token selection.
    \textbf{(b)} Warm single-request decode throughput relative to Full.
    \textbf{(c)} Major decode FLOPs as a percentage of Full.
    Panels (b,c) use Qwen3-1.7B with fixed continuations and prescribed
    Full-call rates; throughput excludes prefill.
    Further details appear in Appendix~\ref{app:efficiency}.}
    \label{fig:main-overview}
    \label{fig:method}
    \label{fig:efficiency}
    \label{fig:system-cost}
\end{figure}

Building on this finding, we introduce On-Demand Attention (ODA), a local-first decoding method that invokes global attention on demand.
Each decoding step begins with local computation, after which a lightweight recall head uses the available decoding states to decide whether to accept the local result or recompute the step with global attention (Figure~\ref{fig:method}(a)).
We train only the recall head with modest data and compute budgets, using the benefit of global attention over local attention under the same history as supervision and keeping pretrained weights unchanged.
ODA also retains the complete historical KV cache, so information skipped at the current step remains available for later access.

ODA performs local computation and makes a recall decision at every step, so its net computational savings depend on whether the avoided global computation outweighs these costs.
Our cost analysis shows that, with a fixed local window and global-attention call frequency, computational savings over full attention increase with context length.
To translate these savings into practical decoding speedups, we build a vLLM~\citep{vllm2023} runtime for on-demand global access that integrates local computation, recall decisions, and conditional global recomputation into GPU-side conditional CUDA Graphs and uses local KV workspace reuse and fused KV commits across layers to reduce synchronization, kernel-launch, and cache-maintenance overhead.

We evaluate ODA at two levels: benefit prediction under fixed histories and recall decisions during free generation.
In fixed-history diagnostics on Qwen3-1.7B~\citep{qwen3_2025}, the recall head captures $2.81\times$ the expected net predictive gain of random selection under the same 40\% offline access budget (Figure~\ref{fig:budget-timing}(a)).
In free-generation evaluations on five RULER16K~\citep{ruler2024} tasks with this model, ODA achieves performance comparable to full attention at a global-attention call frequency of approximately 41\% and outperforms random recall by 56.2 points at similar call frequencies (Figure~\ref{fig:budget-timing}(b)).
These results show that ODA's quality recovery depends on both the frequency of global-attention calls and how those calls are allocated.
Results on RULER at 4K--64K context lengths and evaluations on LongBench~\citep{longbench2024} further support this trade-off between quality and access frequency and demonstrate ODA's applicability across model scales and architectures, including Qwen3, Qwen3.5~\citep{qwen35_2026}, and Gemma~\citep{gemma4_2026} (Tables~\ref{tab:main-results} and~\ref{tab:longbench-main}).
In independent controlled vLLM measurements on Qwen3-1.7B with 512K input tokens and a prescribed 12.5\% global-attention call frequency, ODA achieves $2.65\times$ the single-request decoding throughput of native full attention after warm-up (Figure~\ref{fig:efficiency}(b)).

\begin{samepage}
Our contributions are threefold:
\begin{itemize}
  \item \textbf{Predicting global-attention benefit from pretrained states.}
  We find that the decoding states available after local computation in frozen pretrained models contain signals predictive of the benefit of global attention over local attention, providing empirical support for using existing model representations to guide information access.
  \item \textbf{Local-first On-Demand Attention.}
  We introduce ODA and train only a lightweight recall head to decide when to recompute with global attention, allocating global computation according to decoding needs while keeping pretrained weights unchanged and preserving access to the complete history.
  \item \textbf{Efficient conditional execution.}
  We build a vLLM runtime that combines GPU-side conditional execution with efficient KV-state maintenance and validate the long-context decoding speedups enabled by on-demand access through independent controlled measurements.
\end{itemize}
\end{samepage}

\section{On-Demand Attention}
\label{sec:method}

On-Demand Attention (ODA) gives each Local computation two roles: producing
a candidate state for the current prediction and providing information
for deciding whether global computation is needed. A recall head reads
states available after Local and chooses either to accept the candidate
or to recompute the current step with Full from the same pre-step history
(Figure~\ref{fig:method}(a), lower diagram). ODA trains only the recall head,
keeps pretrained parameters frozen, and retains the complete history so
that positions skipped at one step remain accessible at later steps.

\subsection{Local-first decoding}
\label{sec:local-first}
\label{sec:preserved-history}

At decoding position $t$, the model processes the known input token $x_t$
to predict the next token $x_{t+1}$. Let $C_{t-1}$ denote the state committed
before this step, including historical KV and any other state required by
the model. Local and Full are two access modes of the same frozen backbone
$F_\phi$. Full uses the model's native configuration. Local restricts layers
that natively attend to the full history to $s$ initial positions and a
recent window of width $w$. For these layers, the access sets are
\begin{equation}
\begin{aligned}
    \mathcal S_t^F&=\{1,\ldots,t\},\\
    \mathcal S_t^L&=\{1,\ldots,\min(s,t)\}
    \cup\{\max(1,t-w+1),\ldots,t\}.
\end{aligned}
    \label{eq:access-sets}
\end{equation}
The recent window includes the current token. This initial-token-plus-window
pattern follows StreamingLLM~\citep{streamingllm2024}. For hybrid backbones,
Local restricts only the full-history attention layers; recurrent layers
retain their native computation.

Each step first executes Local and then computes a recall score:
\begin{equation}
\begin{aligned}
    (h_t^L,\Delta C_t^L)&=F_\phi^L(x_t;C_{t-1}),\\
    q_t&=R_\psi\bigl(h_{t-1},E_\phi(x_t),h_t^L\bigr).
\end{aligned}
    \label{eq:recall-head}
\end{equation}
Here $h_t^L$ is the final normalized Local hidden state and $\Delta C_t^L$
is its candidate state update. The head also receives the preceding
step's selected hidden state $h_{t-1}$ and the embedding $E_\phi(x_t)$
of the known input token. All head inputs are available before the current
Full computation and require no reference next token. The head architecture
is detailed in Appendix~\ref{app:head-architecture}.

If $q_t>\theta$, ODA recomputes the current step with Full from the same
$C_{t-1}$ while the Local update remains uncommitted; otherwise, it accepts
Local. The default threshold is $\theta=0$.

Only the selected hidden state $h_t$ passes through the frozen vocabulary
projection $W_{\mathrm{LM}}$ to produce the next-token distribution and
becomes the preceding-state input to the next recall decision. The
corresponding current-position KV and other state updates are committed
to $C_t$. For hybrid backbones, a Full retry also restores the same
pre-step recurrent state.

Algorithm~\ref{alg:oda} gives the complete online step. Decoding starts
from Full prefill, which supplies the initial history, preceding hidden
state, and first generated token (Appendix~\ref{app:prefill-handoff}).

\begin{algorithm}[t]
\caption{One decoding step of On-Demand Attention}
\label{alg:oda}
\small
\begin{algorithmic}[1]
\Require Known input $x_t$, committed history $C_{t-1}$, selected state
$h_{t-1}$, threshold $\theta$
\State $(h_t^L,\Delta C_t^L)\gets F_\phi^L(x_t;C_{t-1})$
\State $q_t\gets R_\psi(h_{t-1},E_\phi(x_t),h_t^L)$
\If{$q_t>\theta$ or $q_t$ is nonfinite}
    \State $(h_t,\Delta C_t)\gets F_\phi^F(x_t;C_{t-1})$
\Else
    \State $(h_t,\Delta C_t)\gets(h_t^L,\Delta C_t^L)$
\EndIf
\State $C_t\gets\Call{Commit}{C_{t-1},\Delta C_t}$
\State $p_t\gets\operatorname{softmax}(W_{\mathrm{LM}}h_t)$
\State $x_{t+1}\gets\Call{Decode}{p_t}$
\State \Return $x_{t+1},C_t,h_t$
\end{algorithmic}
\end{algorithm}

\subsection{Learning to predict recall benefit}
\label{sec:access-value}
\label{sec:recall-head}
\label{sec:head-training}

The recall decision asks how much an additional Full computation would
improve the current prediction. We measure this benefit by comparing
the two branches at the same history and on the same reference next
token $x_{t+1}$. For branch $b\in\{L,F\}$, let $p_t^b$ be its next-token
distribution and $\ell_t^b=-\log p_t^b(x_{t+1})$ its negative
log-likelihood (NLL). The access gain is
\begin{equation}
    g_t=\ell_t^L-\ell_t^F
    =\log p_t^F(x_{t+1})-\log p_t^L(x_{t+1}).
    \label{eq:access-gain}
\end{equation}
Positive gain means that Full increases the reference token's probability;
negative gain favors Local. Even when both branches have the same
top-ranked token, their assigned probabilities can differ. Access gain
therefore captures changes that prediction agreement alone misses, as
illustrated by the schematic Friday example in Figure~\ref{fig:method}(a).

Let $a_t\in\{0,1\}$ denote the access action, with $a_t=0$ accepting Local
and $a_t=1$ invoking Full. Assign the Full call a penalty $\lambda\geq0$
on the prediction-loss scale. At a fixed history and reference token,
the penalized one-step loss is
\begin{equation}
\begin{aligned}
    \ell_t(a_t)
    &=(1-a_t)\ell_t^L+a_t(\ell_t^F+\lambda)\\
    &=\ell_t^L-a_t(g_t-\lambda).
\end{aligned}
    \label{eq:net-gain}
\end{equation}
Invoking Full lowers this loss when its predictive improvement exceeds
the assigned penalty. Relative to the better action, an incorrect choice
incurs an excess loss of $|g_t-\lambda|$
(Appendix~\ref{app:routing-loss}). Two incorrect decisions can therefore
have very different consequences. Retaining gain magnitude in the
supervision distinguishes marginal improvements from high-value
accesses, a distinction that a binary benefit label alone does not
express. This comparison concerns one-step prediction at a fixed history.

To construct paired supervision, we first run the frozen Full backbone
on reference sequences to form the history, then compute Local and Full
candidates from the same pre-step reference history at each position.
These candidates leave the history read by later training positions
unchanged. Training hidden states are selected using reference gains and
shifted by one position within each sequence to form the head's
preceding-state input. During deployment, both the history and the
preceding hidden state follow the head's actual branch selections.
Training thus uses paired supervision on reference histories, while
decoding follows the history formed by its own decisions. The full
construction is given in Appendix~\ref{app:three-stream} and
Algorithm~\ref{alg:oda-training}.

We regress the gain after subtracting the call penalty,
$d_t=g_t-\lambda$. A signed-log transform compresses large target values
while preserving their sign and ordering:
\begin{equation}
    y_t=T(d_t),\qquad
    T(u)=\operatorname{sign}(u)\log(1+|u|).
    \label{eq:gain-transform}
\end{equation}
The head learns this target with Huber regression~\citep{huber1964},
using a transition point of one:
\begin{equation}
    \mathcal L(\psi)=\frac{1}{|\mathcal E|}
    \sum_{t\in\mathcal E}\ell_{\mathrm{Huber},1}(q_t-y_t).
    \label{eq:head-loss}
\end{equation}
Here $\mathcal E$ contains eligible positions in a training batch, each
with a valid reference next token and different Local and Full access
sets. Training updates only the recall-head parameters $\psi$; the
backbone, embedding, and vocabulary projection remain frozen.
Sequence-boundary handling and loss reduction across batches and devices
are detailed in Appendices~\ref{app:training-history}
and~\ref{app:regression-semantics}.

The head learns a regression score $q_t$ from transformed gains for online
branch selection. The penalty $\lambda$ controls the access preference during
training, while $\theta$ sets the decision threshold during deployment.

\subsection{Computational cost and efficient execution}
\label{sec:oda-runtime}
\label{sec:oda-cost}

\paragraph{Computational savings.}
Let $F_{\mathrm{Local}}$ and $F_{\mathrm{Full}}(n)$ denote the two
branches' backbone FLOPs at history length $n$, excluding the vocabulary
projection, and let $F_{\mathrm{Recall}}$ denote the recall-head FLOPs.
At a Full-call rate $\rho$, ODA's average per-step savings relative to
Full are
\begin{equation}
    \Delta F(n,\rho)
    =(1-\rho)F_{\mathrm{Full}}(n)
    -F_{\mathrm{Local}}-F_{\mathrm{Recall}}.
    \label{eq:oda-cost}
\end{equation}
Net savings depend on whether the skipped Full computation covers the
Local and recall costs paid at every step, including recalled steps.
Both methods project the final selected hidden state to the vocabulary
once, so this common cost cancels in the difference. With a fixed Local range and
call rate $\rho<1$, these savings increase with context length.

\paragraph{Efficient execution.}
To translate these computation savings into faster decoding, our Qwen3
runtime in vLLM~\citep{vllm2023} provides an execution path for Local attempts
and conditional Full recomputation. Full retains native FlashAttention
and the paged KV cache. Local uses a dedicated window-restricted attention
path with a persistent ring KV workspace to bound KV reads and avoid
repeated window materialization. GPU-side conditional CUDA Graphs integrate
Local, recall-head scoring, conditional Full, and selected-state commitment,
so branch selection proceeds without copying each score to the CPU for
scheduling. Fused cache updates across layers and graph reuse within
compatible length intervals further reduce small-kernel launch and graph
management overhead. These optimizations preserve the recall head's actual
computation, absolute token positions for Local access, and selected-history
commitment semantics. Cache layout, graph reuse, and the complete latency
model are detailed in
Appendices~\ref{app:runtime-implementation} and~\ref{app:speed-model}.

\section{Experiments}
\label{sec:experiments}

We evaluate ODA's quality--access trade-off across models, tasks, and context
lengths. Benefit diagnostics, free-generation controls, and runtime measurements
examine the role of access selection and its execution benefits.

\subsection{Experimental setup}
\label{sec:experiment-setup}

\paragraph{Tasks and models.}
We evaluate on RULER and LongBench v1~\citep{ruler2024,longbench2024}.
RULER tests controlled long-context capabilities, including retrieval,
tracing, and aggregation, at 4K, 8K, 16K, 32K, and 64K tokens. Our LongBench
evaluation comprises ten tasks spanning single-document QA, multi-document
QA, summarization, and passage retrieval; throughout this paper, LongBench
refers to this task set. Qwen3-1.7B is our primary model. We additionally
evaluate Qwen3-8B~\citep{qwen3_2025}, Qwen3.5-2B,
Qwen3.5-27B~\citep{qwen35_2026}, and Gemma-4-12B-it~\citep{gemma4_2026}
to examine applicability across model scales, families, and native architectures.

\paragraph{Compared policies and shared settings.}
We compare ODA, each model's native attention configuration (Full), and a
fixed local-access policy using the StreamingLLM pattern~\citep{streamingllm2024}.
StreamingLLM and ODA's Local branch share the same access range: the first four
positions and the most recent 2,048 positions, including the current token.
ODA's recall head decides whether to perform additional Full computation.
In hybrid architectures, the local restriction applies only to native global
attention layers; all other layers retain their native computation.
All generation evaluations use Full prefill, after which each policy generates
its own continuation and maintains its own history. We train a separate recall
head for each model while keeping the pretrained backbone frozen.

\paragraph{Metrics and evaluation protocol.}
We use each task's official quality metric and report task-macro scores and
actual Full-call rates during decoding. Table captions specify aggregation
across tasks and context lengths. Training configurations, checkpoints,
thresholds, decoding settings, and metric definitions appear in
Appendix~\ref{app:evaluation-details}; experiment-specific conditions are
given in the corresponding sections.

\subsection{Generation quality under selective global access}
\label{sec:main-results}
\label{sec:hybrid-results}

\begin{table}[t]
\caption{\textbf{RULER generation quality by context length.}
The percentage below each ODA score is its realized Full-call rate,
pooled over routed decoding positions at that length.
Avg takes the unweighted mean over the five lengths for scores and call rates separately.}
\label{tab:main-results}
\label{tab:qwen3-main}
\centering
{\fontsize{7}{8}\selectfont
\newcommand{\odascorewithrate}[2]{\shortstack[c]{#1\\[0pt]{\fontsize{6}{7}\selectfont\textcolor{odaRate}{#2\%}}}}
\setlength{\tabcolsep}{4.5pt}
\begin{tabular}{@{}ll*{5}{c}!{\hspace{3pt}\vrule width 0.4pt\hspace{3pt}}c@{}}
\toprule
\textbf{Model} & \textbf{Method} & \textbf{4k} & \textbf{8k} & \textbf{16k} & \textbf{32k} & \textbf{64k} & \textbf{Avg} \\
\midrule
 & Full-attn & 90.63 & 87.10 & 81.94 & 75.63 & 64.61 & 79.98 \\
 & StreamingLLM & 61.29 & 33.51 & 19.23 & 12.90 & 9.11 & 27.21 \\
\rowcolor{odaRow}
\cellcolor{white}\multirow{-3}{*}{Qwen3-1.7B} & \textbf{ODA} & \odascorewithrate{\textbf{90.12}}{54.69} & \odascorewithrate{\textbf{86.57}}{60.76} & \odascorewithrate{\textbf{81.17}}{41.64} & \odascorewithrate{\textbf{74.68}}{65.45} & \odascorewithrate{\textbf{63.90}}{69.63} & \odascorewithrate{\textbf{79.29}}{58.43} \\
\midrule
 & Full-attn & 95.19 & 94.45 & 92.98 & 91.52 & 83.11 & 91.45 \\
 & StreamingLLM & 67.23 & 40.37 & 25.45 & 17.08 & 14.22 & 32.87 \\
\rowcolor{odaRow}
\cellcolor{white}\multirow{-3}{*}{Qwen3-8B} & \textbf{ODA} & \odascorewithrate{\textbf{95.02}}{40.75} & \odascorewithrate{\textbf{93.79}}{48.40} & \odascorewithrate{\textbf{92.05}}{54.17} & \odascorewithrate{\textbf{90.63}}{57.55} & \odascorewithrate{\textbf{82.13}}{57.87} & \odascorewithrate{\textbf{90.72}}{51.75} \\
\midrule
 & Full-attn & 94.54 & 94.62 & 94.31 & 93.99 & 93.10 & 94.11 \\
 & StreamingLLM & 67.59 & 40.83 & 28.27 & 19.88 & 13.64 & 34.04 \\
\rowcolor{odaRow}
\cellcolor{white}\multirow{-3}{*}{Qwen3.5-2B} & \textbf{ODA} & \odascorewithrate{\textbf{93.82}}{34.33} & \odascorewithrate{\textbf{93.91}}{39.50} & \odascorewithrate{\textbf{94.07}}{44.06} & \odascorewithrate{\textbf{93.58}}{46.77} & \odascorewithrate{\textbf{92.24}}{47.01} & \odascorewithrate{\textbf{93.52}}{42.33} \\
\midrule
 & Full-attn & 97.69 & 92.09 & 95.91 & 97.36 & 97.21 & 96.05 \\
 & StreamingLLM & 66.42 & 45.56 & 31.36 & 25.60 & 20.97 & 37.98 \\
\rowcolor{odaRow}
\cellcolor{white}\multirow{-3}{*}{Qwen3.5-27B} & \textbf{ODA} & \odascorewithrate{\textbf{95.98}}{27.52} & \odascorewithrate{\textbf{90.15}}{35.25} & \odascorewithrate{\textbf{95.79}}{29.47} & \odascorewithrate{\textbf{96.23}}{32.59} & \odascorewithrate{\textbf{96.36}}{34.64} & \odascorewithrate{\textbf{94.90}}{31.89} \\
\midrule
 & Full-attn & 97.41 & 97.19 & 96.61 & 96.34 & 94.86 & 96.48 \\
 & StreamingLLM & 67.18 & 41.00 & 30.07 & 22.38 & 17.61 & 35.65 \\
\rowcolor{odaRow}
\cellcolor{white}\multirow{-3}{*}{Gemma-4-12B-it} & \textbf{ODA} & \odascorewithrate{\textbf{97.20}}{61.22} & \odascorewithrate{\textbf{96.72}}{65.87} & \odascorewithrate{\textbf{96.12}}{67.50} & \odascorewithrate{\textbf{95.67}}{68.88} & \odascorewithrate{\textbf{94.02}}{70.91} & \odascorewithrate{\textbf{95.95}}{66.88} \\
\bottomrule
\end{tabular}
\par}
\end{table}

\paragraph{Context length.}
ODA preserves generation quality close to Full across 4K--64K
(Table~\ref{tab:main-results}). As RULER retrieval inputs grow, targets are
more likely to lie outside the fixed local window, increasing the need to
retrieve distant evidence. When the input doubles from 32K to 64K, the
Full-call rate increases by only 0.24--4.18 percentage points across the five
models, with all score gaps to Full below one point at 64K, while Local
continues to degrade. This change is consistent with increased demand for
remote access: ODA compensates for local restrictions through a modest increase
in global calls, while retaining many generation steps completed by Local.

\begin{table}[t]
\caption{\textbf{LongBench v1 generation quality on ten tasks.}
Avg is the unweighted mean of task scores; parentheses report the mean
Full-call rate over the same tasks.}
\label{tab:longbench-main}
\centering
\scriptsize
\renewcommand{\odaavgrate}[1]{{\fontsize{6}{7}\selectfont\textcolor{odaRate}{\, (#1\%)}}}
\setlength{\tabcolsep}{2.4pt}
\begin{tabular}{@{}ll*{10}{c}!{\hspace{1pt}\vrule width 0.4pt\hspace{1pt}}c@{}}
\toprule
\multirow{2}{*}{\textbf{Model}} & \multirow{2}{*}{\textbf{Method}} & \multicolumn{3}{c}{\textbf{Single-doc QA}} & \multicolumn{4}{c}{\textbf{Multi-doc QA}} & \multicolumn{2}{c}{\textbf{Summarization}} & \multicolumn{1}{c}{\shortstack{\textbf{Synthetic}\\\textbf{retrieval}}} & \multirow{2}{*}{\textbf{Avg}} \\
\cmidrule(lr){3-5} \cmidrule(lr){6-9} \cmidrule(lr){10-11} \cmidrule(lr){12-12}
 & & Qasper & MF-En & MF-Zh & 2Wiki & Hotpot & MuSiQue & DuReader & QMSum & VCSum & PRet-Zh & \\
\midrule
 & Full-attn & 37.34 & 46.78 & 55.29 & 34.62 & 39.60 & 16.06 & 32.42 & 22.76 & 15.15 & 95.00 & 39.50 \\
 & StreamingLLM & 31.69 & 30.42 & 39.39 & 29.46 & 31.10 & 8.36 & 17.75 & 20.92 & 13.52 & 48.00 & 27.06 \\
\rowcolor{odaRow}
\cellcolor{white}\multirow{-3}{*}{\shortstack[l]{Qwen3\\1.7B}} & \textbf{ODA} & \textbf{36.84} & \textbf{45.67} & \textbf{54.99} & \textbf{34.59} & \textbf{40.00} & \textbf{16.28} & \textbf{31.61} & \textbf{22.47} & \textbf{14.87} & \textbf{94.67} & \textbf{39.20}\odaavgrate{80.12} \\
\midrule
 & Full-attn & 46.08 & 54.13 & 62.55 & 40.79 & 56.76 & 32.08 & 26.78 & 24.19 & 13.94 & 97.50 & 45.48 \\
 & StreamingLLM & 37.59 & 37.30 & 45.74 & 33.55 & 45.91 & 21.67 & 17.68 & 21.53 & \textbf{14.20} & 49.50 & 32.47 \\
\rowcolor{odaRow}
\cellcolor{white}\multirow{-3}{*}{\shortstack[l]{Qwen3\\8B}} & \textbf{ODA} & \textbf{45.96} & \textbf{53.83} & \textbf{62.55} & \textbf{40.78} & \textbf{56.77} & \textbf{31.89} & \textbf{26.62} & \textbf{24.03} & 14.05 & \textbf{97.50} & \textbf{45.40}\odaavgrate{88.35} \\
\midrule
 & Full-attn & 39.63 & 52.06 & 61.61 & 34.63 & 48.15 & 31.92 & 27.97 & 22.07 & 13.13 & 84.50 & 41.57 \\
 & StreamingLLM & 32.56 & 34.55 & 44.04 & 29.98 & 36.21 & 17.64 & 16.17 & 19.95 & 11.87 & 50.50 & 29.35 \\
\rowcolor{odaRow}
\cellcolor{white}\multirow{-3}{*}{\shortstack[l]{Qwen3.5\\2B}} & \textbf{ODA} & \textbf{39.21} & \textbf{51.75} & \textbf{61.72} & \textbf{34.63} & \textbf{48.14} & \textbf{32.29} & \textbf{28.65} & \textbf{21.87} & \textbf{12.73} & \textbf{84.50} & \textbf{41.55}\odaavgrate{82.23} \\
\midrule
 & Full-attn & 47.63 & 55.24 & 64.73 & 66.79 & 67.58 & 55.71 & 24.07 & 22.23 & 13.11 & 100.00 & 51.71 \\
 & StreamingLLM & 41.36 & 40.28 & 53.37 & 64.01 & 58.73 & 47.17 & 16.27 & 19.72 & 12.45 & 98.00 & 45.14 \\
\rowcolor{odaRow}
\cellcolor{white}\multirow{-3}{*}{\shortstack[l]{Qwen3.5\\27B}} & \textbf{ODA} & \textbf{47.18} & \textbf{55.02} & \textbf{64.89} & \textbf{66.70} & \textbf{67.65} & \textbf{55.75} & \textbf{23.29} & \textbf{22.13} & \textbf{12.94} & \textbf{100.00} & \textbf{51.55}\odaavgrate{53.07} \\
\midrule
 & Full-attn & 48.06 & 58.14 & 66.62 & 65.50 & 63.70 & 46.59 & 27.18 & 24.22 & 14.72 & 100.00 & 51.47 \\
 & StreamingLLM & 41.81 & 42.83 & 51.71 & 60.68 & 54.02 & 33.47 & 16.68 & 20.85 & 13.45 & 98.50 & 43.40 \\
\rowcolor{odaRow}
\cellcolor{white}\multirow{-3}{*}{\shortstack[l]{Gemma-4\\12B-it}} & \textbf{ODA} & \textbf{48.23} & \textbf{56.63} & \textbf{65.61} & \textbf{65.68} & \textbf{63.74} & \textbf{46.01} & \textbf{25.87} & \textbf{24.14} & \textbf{14.26} & \textbf{99.50} & \textbf{50.97}\odaavgrate{58.65} \\
\bottomrule
\end{tabular}

\end{table}

\begin{figure}[t]
\centering
\includegraphics[width=0.9\linewidth]{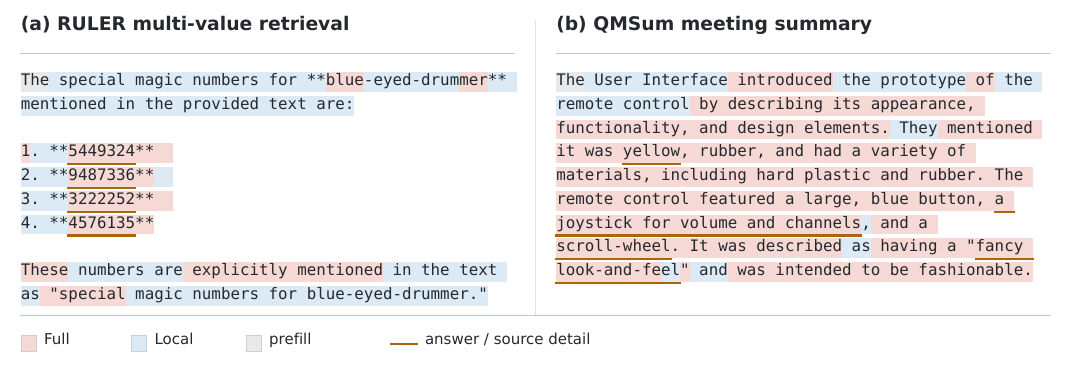}
\caption{\textbf{Token-level access allocation across tasks.}
Routing trajectories from the same Qwen3-1.7B recall head on
\textbf{(a)} RULER multi-value retrieval and \textbf{(b)} an excerpt from a
QMSum meeting summary.}
\label{fig:generation-trajectories}
\end{figure}

\paragraph{Task objectives.}
On LongBench, ODA's average score is within approximately 0.5 points of Full
for all five models (Table~\ref{tab:longbench-main}). Generation trajectories
further illustrate the relationship between task objectives and access
allocation (Figure~\ref{fig:generation-trajectories}). In RULER multi-value
retrieval, Full calls concentrate on the target numbers, while Local handles
much of the introductory text and list formatting. In QMSum, Full calls more
continuously cover descriptions of source details. These examples organize
their outputs around discrete answers and continuous content integration,
respectively, with corresponding differences in the positions and density
of global access.
Additional RULER and MultiNews examples appear in
Appendix~\ref{app:generation-trajectories}.

\paragraph{Model architecture and scale.}
ODA complements models' native history-processing mechanisms, reducing global
calls in full-attention, linear--global hybrid, and window--global hybrid
architectures. On RULER, Qwen3.5 uses fewer Full calls than Qwen3 and Gemma.
We hypothesize that Qwen3.5's recurrent states, which continually aggregate
history, may reduce the need for explicit global reads, whereas Gemma's
windowed computation may depend more on global layers to supply distant
information. Scaling within a family also suggests potential for further
access savings: increasing Qwen3 from 1.7B to 8B and Qwen3.5 from 2B to 27B
reduces the average Full-call rate by 6.68 and 10.44 percentage points,
respectively, while each model's average score remains within 1.2 points of
its own Full baseline. The larger Qwen3.5 model similarly preserves near-Full
quality with fewer calls on LongBench. These results suggest that architectural
pathways for historical information and greater representational capacity
may offer further opportunities to improve ODA's quality--access trade-off.

\subsection{Effectiveness of state-dependent access}
\label{sec:state-dependent-access}
\label{sec:state-ablation}
\label{sec:timing-value}

\begin{figure}[t]
\centering
\includegraphics[width=0.9\linewidth]{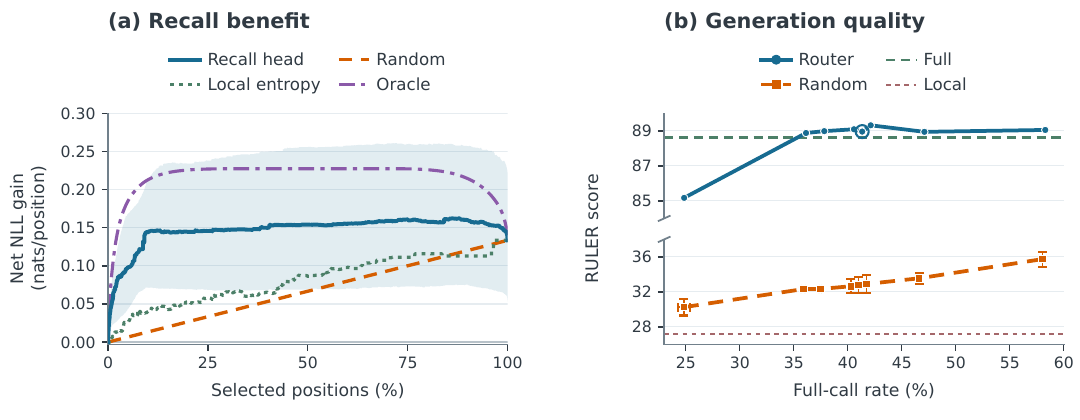}
\caption{\textbf{Global-access benefit and generation quality (Qwen3-1.7B).}
\textbf{(a)} Selected signed NLL gains, normalized by all reference positions.
\textbf{(b)} Quality versus Full-call rate, averaged over five RULER16K tasks.
Shading: 95\% sequence-bootstrap intervals; Random error bars: standard
deviations across three seeds. The ring marks $\theta=0$.}
\label{fig:budget-timing}
\label{fig:recall-budget}
\end{figure}

\paragraph{Access selection.}
Figure~\ref{fig:budget-timing} compares state-dependent access with random
calls through predictive benefit and free generation. On Qwen3-1.7B's fixed
reference records, all rankings use the same histories and selection budgets.
At a 40\% budget, the recall head captures 0.150 nats of net gain per reference
position, $2.81\times$ the expected gain of random selection and above Local
entropy's 0.065. The difference is more pronounced in free generation: on five
RULER16K tasks, ODA scores 88.96 at approximately 41\% Full calls, comparable
to Full, whereas Random scores only 32.79 at a similar call rate. These
comparisons show that using model states to select access positions both
captures more predictive benefit within a fixed budget and yields better
generation quality at a similar proportion of global calls.

\paragraph{Benefit concentration.}
The learned-policy curves show large early gains followed by smaller increments.
In Figure~\ref{fig:budget-timing}(a), the highest-scoring approximately 10\%
of positions already capture 0.144 nats of net gain per reference position;
expanding the budget to 40\% increases this to only 0.150.
In Figure~\ref{fig:budget-timing}(b), ODA already scores 85.17 at 24.8\%
Full calls, reaches 88.88 at 36.1\%, and stays around 89 across the remaining
measured configurations up to 58.3\%. In comparison, random selection's
expected benefit grows linearly with its budget, while random calling yields
only gradual gains in generation score. These patterns indicate that access
value varies across positions: the learned policy prioritizes beneficial
accesses and recovers much of the quality at relatively low call rates,
with smaller increments from additional calls over these measured ranges.
This leaves room to retain valuable accesses while assigning more generation
steps to Local.

\paragraph{State inputs.}
The current Local computation provides evidence for the recall decision.
Under shared training settings and a zero decision threshold
(Table~\ref{tab:state-ablation}), using only the preceding selected hidden
state $H$ and current token embedding $E$ yields 81.94, but invokes Full on
99.73\% of steps. Adding the current Local state $L$ maintains a score of
81.57 while reducing the Full-call rate to 76.22\%. Using $L$ alone yields
a similar 81.60 but requires 89.25\% Full calls. The combined inputs provide
the preceding context representation, current token information, and current
Local result, supporting the head in accepting more Local outputs while
preserving similar quality. This agrees with ODA's local-first design:
first form the current prediction state through local computation, then
decide whether additional global information is needed.

\begin{table}[t]
\centering
\caption{\textbf{Effect of recall-head inputs on quality and Full-call rate.}
Qwen3-1.7B heads share the same architecture and are evaluated on RULER16K
at $\theta=0$.
$H$, $E$, and $L$ denote the preceding selected hidden state, current
token embedding, and current Local state, respectively.}
\label{tab:state-ablation}
\small
\begin{tabular}{@{}cccrr@{}}
\toprule
$H$ & $E$ & $L$ & RULER16K & Full (\%) \\
\midrule
\checkmark & --- & --- & 76.84 & 83.20 \\
--- & \checkmark & --- & 49.88 & 87.32 \\
--- & --- & \checkmark & 81.60 & 89.25 \\
\checkmark & \checkmark & --- & 81.94 & 99.73 \\
\checkmark & --- & \checkmark & 67.96 & 21.86 \\
--- & \checkmark & \checkmark & 81.53 & 97.51 \\
\midrule
\checkmark & \checkmark & \checkmark & 81.57 & 76.22 \\
\bottomrule
\end{tabular}

\end{table}

These results indicate further potential for state-dependent access allocation.
Global access provides unequal benefits across generation positions, and
prioritizing high-value accesses allows fewer calls to recover much of the
generation quality. Under fixed histories, the offline Oracle still captures
more benefit than the current recall head, suggesting room to improve this
trade-off through more accurate selection. As context length grows, each
avoided global access also saves more computation
(Section~\ref{sec:oda-cost}). Better identification of access benefit may
therefore further reduce expensive global computation while preserving
long-context generation quality.

\subsection{Computation and decoding efficiency}
\label{sec:decode-throughput}

Using the vLLM runtime in Section~\ref{sec:oda-runtime}, we evaluate
Qwen3-1.7B's computational savings and decode throughput across context
lengths and Full-call rates. At each length, all methods share the input
and a fixed continuation, with Full calls controlled by prescribed
schedules. The recall head executes at every ODA step, and timing covers
the complete decoding path, including conditional recomputation and state
maintenance.

Figure~\ref{fig:efficiency}(c) shows major decoding FLOPs as a percentage
of Full at each context length. For the 512K
input and 12.5\% Full calls, ODA uses 15.69\% of Full's FLOPs,
an 84.31\% reduction and a $6.38\times$ computation ratio. The relative cost
decreases with context length, while larger recall budgets leave more
repeated computation.

Figure~\ref{fig:efficiency}(b) reports the resulting decode throughput.
Measurements use batch size one and CUDA Graphs for both ODA and native
Full, with matched hardware configurations within each length. At the same
512K/12.5\% operating point, median throughput rises from 77.64 to 205.71
tokens/s ($2.65\times$).
At 4K, however, ODA remains 13.1\% slower at the same call rate: avoiding short global reads does
not offset the Local and conditional-execution overhead. Thus, the speed
benefit depends on both context length and recall frequency.

Further experimental details are provided in Appendix~\ref{app:efficiency}.

\section{Related Work}
\label{sec:related-work}

\paragraph{Dynamic local--global attention.}
AHA learns per-token choices between global and sliding-window attention
for each head~\citep{aha2025}, while Switch Attention learns token- and
layer-dependent routing through continued pretraining~\citep{switch2026}.
L2A uses local outputs to gate a global branch within each layer~\citep{l2a2026},
and LoGo couples local and global branches with a controller for the
global-attention budget~\citep{logo2026}. ODA makes a single
recall decision after a complete Local forward pass and trains only the
recall head, keeping the pretrained backbone frozen.

\paragraph{Sparse access and KV management.}
H$_2$O evicts KV entries~\citep{h2o2023}, while QUEST selects pages from
retained KV using the current query~\citep{quest2024}. SeerAttention-R
learns block selection through attention distillation with frozen pretrained
weights~\citep{seerattentionr2025}. RefreshKV alternates full-context and
subset attention, updating a small cache using attention from each global
step~\citep{refreshkv2025}. Like RefreshKV, ODA preserves subsequent
access to the complete history; it learns when to extend a fixed Local
read to Full rather than which remote KV subset to read.

\paragraph{Model-guided information acquisition.}
FLARE uses low-confidence tokens in a draft sentence to trigger retrieval
and regeneration~\citep{flare2023}, while Probing-RAG reads hidden states
to decide whether further document retrieval is needed~\citep{probingrag2025}.
Declarative Attention uses model-generated \texttt{<global>},
\texttt{<focus>}, and \texttt{<local>} declarations to control subsequent attention without
additional training~\citep{da2026}. LongPPL uses long--short context
prediction differences to identify key tokens for evaluation and training
reweighting~\citep{longppl2025}. ODA uses paired Full--Local prediction
gains under a shared history to supervise a state readout that decides
whether to recompute the current step with Full.

\section{Discussion and Conclusion}
\label{sec:discussion}
\label{sec:conclusion}

\paragraph{Discussion.}
ODA retains the complete history for later recall, while KV storage grows
with sequence length. Access decisions shape subsequent states, so later
Full computation uses the
policy's actual history rather than reconstructing an Always-Full trajectory.
The head learns one-step gains on reference histories constructed with Full,
whereas deployment histories evolve with its own decisions. Future work could
learn recall on policy-generated histories and account for downstream
generation quality and cumulative computational cost.

\paragraph{Conclusion.}
We find that states available after local computation in frozen models contain
signals predictive of the benefit of additional global attention.
ODA reads out these signals with a lightweight recall head, recovering most
of the generation quality lost under local attention with fewer global calls
while retaining the complete history.
Our vLLM runtime with GPU-side conditional execution delivers decoding
speedups in independent controlled long-context measurements.
These findings show that pretrained decoding states can support both token
prediction and the on-demand allocation of global computation.

\section*{AI Use Statement}

Generative AI assistance (OpenAI Codex) was used to support manuscript
writing, translation, and language editing. This included suggesting word
choices and alternative phrasing, improving grammar and sentence structure,
and refining wording for clarity, conciseness, and consistency. The authors
take responsibility for the final content of the manuscript.

\section*{Reproducibility Statement}

Algorithms~\ref{alg:oda} and~\ref{alg:oda-training} specify the decoding
procedure and training update.
Appendix~\ref{app:method} details the supervision, state handling, and
computational cost assumptions.
Appendix~\ref{app:evaluation-details} describes the training and evaluation
settings, and Appendix~\ref{app:training-cost} provides the primary
training configuration and resource measurements.
Appendix~\ref{app:experiments} reports per-task results and diagnostic
protocols, including the controlled runtime measurements in
Appendix~\ref{app:efficiency}.
The anonymous supplementary package provides implementation code,
experiment configurations, numerical records, and scripts for checking
aggregates and regenerating plots. Its README documents dependencies,
data organization, and execution commands.

\bibliography{iclr2027_conference}

@article{l2a2026,
  title = {Learning When to Attend: Conditional Memory Access for Long-Context {LLMs}},
  author = {Choudhary, Sakshi and Chattopadhyay, Aditya and Zancato, Luca and Nunez, Elvis and Trager, Matthew and Xia, Wei and Soatto, Stefano},
  journal = {arXiv preprint arXiv:2603.17484},
  year = {2026},
  url = {https://arxiv.org/abs/2603.17484}
}

@article{switch2026,
  title = {Switch Attention: Towards Dynamic and Fine-grained Hybrid Transformers},
  author = {Zhao, Yusheng and Li, Hourun and Wu, Bohan and Yin, Yichun and Shang, Lifeng and Yuan, Jingyang and Zhang, Meng and Zhang, Ming},
  journal = {arXiv preprint arXiv:2603.26380},
  year = {2026},
  url = {https://arxiv.org/abs/2603.26380}
}

@inproceedings{probingrag2025,
  title = {Probing-{RAG}: Self-Probing to Guide Language Models in Selective Document Retrieval},
  author = {Baek, Ingeol and Chang, Hwan and Kim, ByeongJeong and Lee, Jimin and Lee, Hwanhee},
  booktitle = {Findings of the Association for Computational Linguistics: NAACL 2025},
  year = {2025},
  publisher = {Association for Computational Linguistics},
  pages = {3287--3304},
  doi = {10.18653/v1/2025.findings-naacl.181},
  url = {https://aclanthology.org/2025.findings-naacl.181/}
}

@inproceedings{ruler2024,
  title = {{RULER}: What's the Real Context Size of Your Long-Context Language Models?},
  author = {Hsieh, Cheng-Ping and Sun, Simeng and Kriman, Samuel and Acharya, Shantanu and Rekesh, Dima and Jia, Fei and Zhang, Yang and Ginsburg, Boris},
  booktitle = {First Conference on Language Modeling},
  year = {2024},
  url = {https://arxiv.org/abs/2404.06654}
}

@inproceedings{longbench2024,
  title = {{LongBench}: A Bilingual, Multitask Benchmark for Long Context Understanding},
  author = {Bai, Yushi and Lv, Xin and Zhang, Jiajie and Lyu, Hongchang and Tang, Jiankai and Huang, Zhidian and Du, Zhengxiao and Liu, Xiao and Zeng, Aohan and Hou, Lei and Dong, Yuxiao and Tang, Jie and Li, Juanzi},
  booktitle = {Proceedings of the 62nd Annual Meeting of the Association for Computational Linguistics (Volume 1: Long Papers)},
  year = {2024},
  publisher = {Association for Computational Linguistics},
  pages = {3119--3137},
  doi = {10.18653/v1/2024.acl-long.172},
  url = {https://aclanthology.org/2024.acl-long.172/}
}

@article{qwen3_2025,
  title = {{Qwen3} Technical Report},
  author = {{Qwen Team}},
  journal = {arXiv preprint arXiv:2505.09388},
  year = {2025},
  url = {https://arxiv.org/abs/2505.09388}
}

@misc{qwen35_2026,
  title = {{Qwen3.5}: Towards Native Multimodal Agents},
  author = {{Qwen Team}},
  howpublished = {Qwen blog},
  year = {2026},
  month = feb,
  url = {https://qwen.ai/blog?id=qwen3.5}
}

@article{gemma4_2026,
  title = {{Gemma 4} Technical Report},
  author = {{Gemma Team}},
  journal = {arXiv preprint arXiv:2607.02770},
  year = {2026},
  url = {https://arxiv.org/abs/2607.02770}
}

@inproceedings{h2o2023,
  title = {{H$_2$O}: Heavy-Hitter Oracle for Efficient Generative Inference of Large Language Models},
  author = {Zhang, Zhenyu and Sheng, Ying and Zhou, Tianyi and Chen, Tianlong and Zheng, Lianmin and Cai, Ruisi and Song, Zhao and Tian, Yuandong and R{\'e}, Christopher and Barrett, Clark and Wang, Zhangyang and Chen, Beidi},
  booktitle = {Advances in Neural Information Processing Systems},
  volume = {36},
  year = {2023},
  url = {https://proceedings.neurips.cc/paper_files/paper/2023/hash/6ceefa7b15572587b78ecfcebb2827f8-Abstract-Conference.html}
}

@inproceedings{quest2024,
  title = {{QUEST}: Query-Aware Sparsity for Efficient Long-Context {LLM} Inference},
  author = {Tang, Jiaming and Zhao, Yilong and Zhu, Kan and Xiao, Guangxuan and Kasikci, Baris and Han, Song},
  booktitle = {Proceedings of the 41st International Conference on Machine Learning},
  series = {Proceedings of Machine Learning Research},
  volume = {235},
  pages = {47901--47911},
  publisher = {PMLR},
  year = {2024},
  url = {https://proceedings.mlr.press/v235/tang24l.html}
}

@article{da2026,
  title = {Language Models Can Control Their Own Attention},
  author = {Ho, Namgyu and Ahmad, Huzama and Koh, Woosung and Yun, Se-Young and Schuster, Tal and {Nogueira dos Santos}, Cicero},
  journal = {arXiv preprint arXiv:2609.02737},
  year = {2026},
  url = {https://arxiv.org/abs/2609.02737}
}

@misc{anthropic2026cadences,
  title = {{Anthropic Economic Index} Report: Cadences},
  author = {Massenkoff, Maxim and Lyubich, Eva and Sacher, Szymon and Hitzig, Zoe and Zhang, Shaoyi and Heller, Ryan and McCrory, Peter},
  howpublished = {Anthropic research report},
  year = {2026},
  month = jun,
  note = {June 26, 2026},
  url = {https://www.anthropic.com/research/economic-index-june-2026-report}
}

@inproceedings{longspec2026,
  title = {{LongSpec}: Long-Context Lossless Speculative Decoding with Efficient Drafting and Verification},
  author = {Yang, Penghui and Du, Cunxiao and Zhang, Fengzhuo and Wang, Haonan and Pang, Tianyu and Du, Chao and An, Bo},
  booktitle = {Proceedings of the 64th Annual Meeting of the Association for Computational Linguistics (Volume 1: Long Papers)},
  year = {2026},
  publisher = {Association for Computational Linguistics},
  pages = {1826--1844},
  doi = {10.18653/v1/2026.acl-long.83},
  url = {https://aclanthology.org/2026.acl-long.83/}
}

@inproceedings{streamingllm2024,
  title = {Efficient Streaming Language Models with Attention Sinks},
  author = {Xiao, Guangxuan and Tian, Yuandong and Chen, Beidi and Han, Song and Lewis, Mike},
  booktitle = {International Conference on Learning Representations},
  pages = {21875--21895},
  year = {2024},
  url = {https://proceedings.iclr.cc/paper_files/paper/2024/hash/5e5fd18f863cbe6d8ae392a93fd271c9-Abstract-Conference.html}
}

@inproceedings{flare2023,
  title = {Active Retrieval Augmented Generation},
  author = {Jiang, Zhengbao and Xu, Frank and Gao, Luyu and Sun, Zhiqing and Liu, Qian and Dwivedi-Yu, Jane and Yang, Yiming and Callan, Jamie and Neubig, Graham},
  booktitle = {Proceedings of the 2023 Conference on Empirical Methods in Natural Language Processing},
  pages = {7969--7992},
  publisher = {Association for Computational Linguistics},
  year = {2023},
  doi = {10.18653/v1/2023.emnlp-main.495},
  url = {https://aclanthology.org/2023.emnlp-main.495/}
}

@inproceedings{longppl2025,
  title = {What is Wrong with Perplexity for Long-context Language Modeling?},
  author = {Fang, Lizhe and Wang, Yifei and Liu, Zhaoyang and Zhang, Chenheng and Jegelka, Stefanie and Gao, Jinyang and Ding, Bolin and Wang, Yisen},
  booktitle = {International Conference on Learning Representations},
  pages = {94541--94563},
  year = {2025},
  url = {https://proceedings.iclr.cc/paper_files/paper/2025/hash/ebd6641c32ed633c2a3addc689d39896-Abstract-Conference.html}
}

@inproceedings{refreshkv2025,
  title = {{RefreshKV}: Updating Small {KV} Cache During Long-form Generation},
  author = {Xu, Fangyuan and Goyal, Tanya and Choi, Eunsol},
  booktitle = {Proceedings of the 63rd Annual Meeting of the Association for Computational Linguistics (Volume 1: Long Papers)},
  pages = {24878--24893},
  publisher = {Association for Computational Linguistics},
  year = {2025},
  doi = {10.18653/v1/2025.acl-long.1211},
  url = {https://aclanthology.org/2025.acl-long.1211/}
}

@article{aha2025,
  title = {Learning When Not to Attend Globally},
  author = {Luo, Xuan and Zhang, Kailai and Yan, Xifeng},
  journal = {arXiv preprint arXiv:2512.22562},
  year = {2025},
  url = {https://arxiv.org/abs/2512.22562}
}

@article{logo2026,
  title = {{LoGo}: Token-Level Dynamic Local-Global Attention},
  author = {Pan, Yuqi and Li, Zheng and Tang, Bohao and Qin, Zhen and Li, Guoqi},
  journal = {arXiv preprint arXiv:2608.29539},
  year = {2026},
  url = {https://arxiv.org/abs/2608.29539}
}

@article{seerattentionr2025,
  title = {{SeerAttention-R}: Sparse Attention Adaptation for Long Reasoning},
  author = {Gao, Yizhao and Guo, Shuming and Cao, Shijie and Xia, Yuqing and Cheng, Yu and Wang, Lei and Ma, Lingxiao and Sun, Yutao and Ye, Tianzhu and Dong, Li and So, Hayden Kwok-Hay and Hua, Yu and Cao, Ting and Yang, Fan and Yang, Mao},
  journal = {arXiv preprint arXiv:2506.08889},
  year = {2025},
  url = {https://arxiv.org/abs/2506.08889}
}

@article{shortcontextdominance2026,
  title = {Short-Context Dominance: How Much Local Context Natural Language Actually Needs?},
  author = {Vakilian, Vala and Wang, Zimeng and Rawat, Ankit Singh and Thrampoulidis, Christos},
  journal = {arXiv preprint arXiv:2512.08082},
  year = {2026},
  url = {https://arxiv.org/abs/2512.08082}
}

@inproceedings{vllm2023,
  title = {Efficient Memory Management for Large Language Model Serving with {PagedAttention}},
  author = {Kwon, Woosuk and Li, Zhuohan and Zhuang, Siyuan and Sheng, Ying and Zheng, Lianmin and Yu, Cody Hao and Gonzalez, Joseph E. and Zhang, Hao and Stoica, Ion},
  booktitle = {Proceedings of the 29th Symposium on Operating Systems Principles},
  publisher = {Association for Computing Machinery},
  pages = {611--626},
  year = {2023},
  doi = {10.1145/3600006.3613165},
  url = {https://arxiv.org/abs/2309.06180}
}

@article{huber1964,
  title = {Robust Estimation of a Location Parameter},
  author = {Huber, Peter J.},
  journal = {The Annals of Mathematical Statistics},
  volume = {35},
  number = {1},
  pages = {73--101},
  year = {1964},
  doi = {10.1214/aoms/1177703732},
  url = {https://doi.org/10.1214/aoms/1177703732}
}

@inproceedings{rmsnorm2019,
  title = {Root Mean Square Layer Normalization},
  author = {Zhang, Biao and Sennrich, Rico},
  booktitle = {Advances in Neural Information Processing Systems},
  volume = {32},
  year = {2019},
  url = {https://arxiv.org/abs/1910.07467}
}

@article{shazeer2020glu,
  title = {{GLU} Variants Improve Transformer},
  author = {Shazeer, Noam},
  journal = {arXiv preprint arXiv:2002.05202},
  year = {2020},
  url = {https://arxiv.org/abs/2002.05202}
}

@inproceedings{adamw2019,
  title = {Decoupled Weight Decay Regularization},
  author = {Loshchilov, Ilya and Hutter, Frank},
  booktitle = {International Conference on Learning Representations},
  year = {2019},
  url = {https://arxiv.org/abs/1711.05101}
}

@misc{step35sft2026,
  title = {{Step-3.5-Flash-SFT}},
  author = {{StepFun}},
  howpublished = {Hugging Face dataset},
  year = {2026},
  url = {https://huggingface.co/datasets/stepfun-ai/Step-3.5-Flash-SFT}
}

@inproceedings{yarn2024,
  title = {{YaRN}: Efficient Context Window Extension of Large Language Models},
  author = {Peng, Bowen and Quesnelle, Jeffrey and Fan, Honglu and Shippole, Enrico},
  booktitle = {International Conference on Learning Representations},
  year = {2024},
  url = {https://arxiv.org/abs/2309.00071}
}

@inproceedings{zeroscrolls2023,
  title = {{ZeroSCROLLS}: A Zero-Shot Benchmark for Long Text Understanding},
  author = {Shaham, Uri and Ivgi, Maor and Efrat, Avia and Berant, Jonathan and Levy, Omer},
  booktitle = {Findings of the Association for Computational Linguistics: EMNLP 2023},
  publisher = {Association for Computational Linguistics},
  pages = {7977--7989},
  year = {2023},
  doi = {10.18653/v1/2023.findings-emnlp.536},
  url = {https://aclanthology.org/2023.findings-emnlp.536/}
}

@inproceedings{longbenchv2_2025,
  title = {{LongBench v2}: Towards Deeper Understanding and Reasoning on Realistic Long-context Multitasks},
  author = {Bai, Yushi and Tu, Shangqing and Zhang, Jiajie and Peng, Hao and Wang, Xiaozhi and Lv, Xin and Cao, Shulin and Xu, Jiazheng and Hou, Lei and Dong, Yuxiao and Tang, Jie and Li, Juanzi},
  booktitle = {Proceedings of the 63rd Annual Meeting of the Association for Computational Linguistics (Volume 1: Long Papers)},
  publisher = {Association for Computational Linguistics},
  pages = {3639--3664},
  year = {2025},
  doi = {10.18653/v1/2025.acl-long.183},
  url = {https://aclanthology.org/2025.acl-long.183/}
}

@inproceedings{loogle2024,
  title = {{LooGLE}: Can Long-Context Language Models Understand Long Contexts?},
  author = {Li, Jiaqi and Wang, Mengmeng and Zheng, Zilong and Zhang, Muhan},
  booktitle = {Proceedings of the 62nd Annual Meeting of the Association for Computational Linguistics (Volume 1: Long Papers)},
  publisher = {Association for Computational Linguistics},
  pages = {16304--16333},
  year = {2024},
  doi = {10.18653/v1/2024.acl-long.859},
  url = {https://aclanthology.org/2024.acl-long.859/}
}
\bibliographystyle{iclr2027_conference}

\clearpage
\appendix
\section{Additional Method Details}
\label{app:method}

\subsection{Access gain and the cost of routing errors}
\label{app:routing-loss}

At a fixed history and reference token, write
$\ell_t^b=-\log p_t^b(x_{t+1})$ for $b\in\{L,F\}$ and
$g_t=\ell_t^L-\ell_t^F$. Let $a_t\in\{0,1\}$ indicate whether Full is
selected. The excess loss relative to the better current branch is
\begin{equation}
    (1-a_t)\ell_t^L+a_t\ell_t^F-\min(\ell_t^L,\ell_t^F)
    =(1-a_t)[g_t]_+ + a_t[-g_t]_+,
    \label{eq:routing-excess-loss}
\end{equation}
where $[u]_+=\max(u,0)$. Selecting the worse branch costs $|g_t|$, so
routing-error counts alone do not measure the resulting NLL penalty.
With the Full-call penalty $\lambda$, the same identity holds for penalized
losses after replacing $g_t$ by $g_t-\lambda$. This is a comparison on fixed
states: summing it does not give the loss of an alternative generated
trajectory. Huber regression is a surrogate and does not directly minimize
this excess loss.

\paragraph{Reference benefit and distribution disagreement.}
KL divergence $D_{\mathrm{KL}}(p_t^F\Vert p_t^L)$ averages
$\log p_t^F(Y)-\log p_t^L(Y)$ over $Y\sim p_t^F$.
It therefore measures expected gain under Full's predictive distribution,
whereas $g_t$ evaluates the reference next token. Exact current-step KL
also requires executing Full, so it is an offline diagnostic.

\subsection{Recall-head architecture}
\label{app:head-architecture}

For Qwen3-1.7B, each of the three 2,048-dimensional inputs passes through
an independent RMSNorm~\citep{rmsnorm2019}, a biased
$2048\!\to\!1024$ projection, and SiLU.
For tower outputs $a,b,c\in\mathbb R^{1024}$, the interaction vector is
\begin{equation}
    u=[a;b;c;a\odot b;|a-b|;a\odot c;|a-c|;b\odot c;|b-c|]
    \in\mathbb R^{9216}.
\end{equation}
A biased fusion projection gives
\begin{equation}
    z_0=\tfrac12\bigl[a+b+c+\operatorname{SiLU}(W_fu+b_f)\bigr],
    \label{eq:head-fusion}
\end{equation}
with $W_f\in\mathbb R^{1024\times9216}$. Two pre-RMSNorm residual
SwiGLU blocks~\citep{shazeer2020glu} follow, each with bias-free $1024\!\to\!2048$ gate and up
projections and a $2048\!\to\!1024$ down projection. A final RMSNorm and
biased scalar projection produce $q_t$. The head has 28,325,889 parameters
and uses FP32 weights and arithmetic. The backbone, embedding, and
vocabulary projection remain frozen. Head training uses neither dropout
nor weight decay.

\subsection{Paired supervision and historical inputs}
\label{app:three-stream}
\label{app:local-mask}
\label{app:training-history}

\paragraph{Branch computation.}
The access sets are defined in Equation~\ref{eq:access-sets}, with $s=4$
and $w=2048$. They coincide through position 2,052; eligible supervision
begins at position 2,053. Padding and packed-sequence boundaries prevent
cross-sequence attention. For either branch,
\begin{equation}
    (h_t^b,\Delta C_t^b)=F_\phi^b(x_t;C_{t-1}),\qquad b\in\{L,F\},
    \label{eq:branch}
\end{equation}
where $h_t^b$ is the final normalized hidden state. The candidate update
$\Delta C_t^b$ contains current-position KV at every attention layer and,
for hybrid backbones, the native recurrent-state updates.

Training computes a Full trunk and two current-position counterfactuals.
Both counterfactuals read the trunk's historical state, while propagating
their own current-position activations through the layers. Past attention
uses $j<t$; each candidate's current KV contribution is merged separately
by online softmax. Neither counterfactual changes the trunk history used
at later positions. The Full counterfactual and trunk are mathematically
equivalent in exact arithmetic, but their different BF16 reduction paths
can change small gain signs. The reported implementation therefore uses
the separate Full counterfactual for the Full NLL target.

The frozen vocabulary projection supplies both reference-token NLLs.
Vocabulary log-sum-exp is evaluated in chunks of 16 positions to avoid
storing a sequence-by-vocabulary tensor. Eligibility requires valid current
and next tokens, an unmasked reference label, consecutive position IDs
within the same segment, and context length greater than 2,052. Backbone
features and labels are computed without gradients.

\paragraph{Previous hidden state during training.}
With eligibility indicator $m_t$, the preceding-state input is constructed
from the selected training states
\begin{equation}
    h_t^{\mathrm{train}}=
    \begin{cases}
        h_t^{L,\mathrm{cf}}, & m_t=1\ \text{and}\ g_t<\tau_{\mathrm{hist}},\\
        h_t^{F,\mathrm{trunk}}, & \text{otherwise}.
    \end{cases}
    \label{eq:training-history}
\end{equation}
These states are detached and shifted within each sequence, so the head at
$t$ receives $h_{t-1}^{\mathrm{train}}$, corresponding to $H^-_t$ in
Algorithm~\ref{alg:oda-training}. The input is zero at each segment start.
The Full branch of this feature uses the trunk hidden state; the Full
counterfactual is used for the gain target. This selection does not alter
training KV. Because $g_{t-1}$ is measured on the already observed $x_t$,
it does not depend on the current reference target $x_{t+1}$. It nevertheless
requires both preceding predictions, whereas deployment uses the branch
actually selected by the head. Training and deployment therefore differ
in both historical-state construction and preceding-state selection.

\begin{algorithm}[t]
\caption{One recall-head training update}
\label{alg:oda-training}
\small
\noindent Pretrained parameters $\phi$ are frozen; only $R_\psi$ is trained.
\par\medskip
\begin{algorithmic}[1]
\Require Token batch $X$, Full-call penalty $\lambda$, history threshold
$\tau_{\mathrm{hist}}$
\State $\mathcal E\gets\Call{Eligible}{X}$
\State $(C^F,H^{\mathrm{tr}})\gets\Call{FullTrunk}{X}$
\State $(H^L,H^F)\gets\Call{Pair}{X,C^F}$
\State $g\gets\Call{AccessGain}{H^L,H^F,X}$
\State $H^{\mathrm{train}}\gets\Call{Select}{H^L,H^{\mathrm{tr}},g,\tau_{\mathrm{hist}}}$
\State $H^{-}\gets\Call{Shift}{H^{\mathrm{train}}}$
\For{$t\in\mathcal E$}
    \State $q_t\gets R_\psi(H^{-}_t,E_\phi(x_t),h_t^L)$
    \State $y_t\gets T(g_t-\lambda)$
\EndFor
\State $\mathcal L\gets\Call{HuberReduce}{q,y,\mathcal E}$
\State Update $\psi$ using $\nabla_\psi\mathcal L$
\end{algorithmic}
\medskip
\noindent\textsc{Pair} computes both candidates from the Full-trunk history.
\textsc{Select} uses $H^L$ at eligible positions with
$g_t<\tau_{\mathrm{hist}}$, and $H^{\mathrm{tr}}$ otherwise.
\textsc{Shift} moves the selected states one position forward within each
sequence, using zero at its start.
\textsc{HuberReduce} applies Equation~\ref{eq:head-loss}, using the
configuration-specific reduction and empty-batch handling in
Appendix~\ref{app:regression-semantics}.
\end{algorithm}

\paragraph{Prefill handoff.}
\label{app:prefill-handoff}
For prompt $x_{1:m}$, Full prefill commits $C_m$, sets $h_m=h_m^F$, and
produces the first output $x_{m+1}$. The first routed step consumes this
token with $(C_m,h_m)$ and produces $x_{m+2}$. Consequently, $N$ generated
outputs contain $N-1$ routed steps. All reported routing counts exclude
the output supplied directly by prefill.

\subsection{Regression target and reduction}
\label{app:regression-semantics}

For residual $e=q_t-y_t$, the loss is
\begin{equation}
    \ell_{\mathrm{Huber},1}(e)=
    \begin{cases}
        \tfrac12 e^2,& |e|\leq1,\\
        |e|-\tfrac12,& |e|>1.
    \end{cases}
\end{equation}
All eligible gains, including near-zero values, are retained. There is no
gain-dependent weighting or auxiliary sign-classification loss.
The primary Qwen3-1.7B configuration averages the distributed eligible-token
means of its accumulated microbatches. The input ablations and
Qwen3.5-2B training instead divide the summed losses by the total
eligible-token count over the entire optimizer update. For accumulated
microbatch sets $\mathcal E_j$ pooled across data-parallel workers, these
reductions are, respectively,
\begin{equation}
\begin{aligned}
    \mathcal L_{\mathrm{micro}}
    &=\frac{1}{M}\sum_{j=1}^{M}
      \frac{\sum_{t\in\mathcal E_j}\ell_t}{|\mathcal E_j|},\\
    \mathcal L_{\mathrm{step}}
    &=\frac{\sum_{j=1}^{M}\sum_{t\in\mathcal E_j}\ell_t}
            {\sum_{j=1}^{M}|\mathcal E_j|},
\end{aligned}
\end{equation}
shown for nonempty microbatches. They need not agree when eligible-token
counts differ. In the step-normalized implementation, empty microbatches
contribute zero and an entirely empty update is skipped. Training settings
for the input ablation are recorded in Appendix~\ref{app:local-state-ablation};
model-specific evaluation settings are given in Appendix~\ref{app:evaluation-details}.

The Qwen3-1.7B head used for RULER16K uses $\lambda=0$ and
$\tau_{\mathrm{hist}}=0.001$; its LongBench head uses
$\lambda=0$ and $\tau_{\mathrm{hist}}=0$. The Qwen3.5-2B head and input
ablations use $\lambda=0.001$ and $\tau_{\mathrm{hist}}=0$.
The Full-call penalty is expressed on the NLL scale and is not a measured
hardware cost. The history threshold and deployment score threshold are
separate parameters. Although the signed-log transform is invertible,
Huber regression on a transformed target does not make its inverse an
estimate of the conditional mean raw gain. We use $q_t$ as a regression
score, without interpreting it as a probability or calibrated NLL benefit.

\subsection{Selected-state execution}
\label{app:runtime-implementation}

Let $b_t=F$ when $q_t>\theta$ and $b_t=L$ otherwise; nonfinite scores
fall back to Full. The selected branch determines all committed outputs:
\begin{equation}
\begin{aligned}
    h_t&=h_t^{b_t},&
    p_t&=\operatorname{softmax}(W_{\mathrm{LM}}h_t),\\
    C_t&=C_{t-1}\mathbin{\Vert}\Delta C_t^{b_t}.
\end{aligned}
    \label{eq:oda-commit}
\end{equation}
The operator $\Vert$ appends the selected current KV and applies any
recurrent-state updates. The same $h_t$ becomes the preceding hidden input
at the next step; it is updated after Local as well as Full selections.
Candidate state remains uncommitted until the action is final. Only the
selected hidden passes through the vocabulary projection at deployment.
Full recall reads the history produced by ODA's actual actions, without
reconstructing an Always-Full trajectory.

\paragraph{Complete history and Local workspace.}
The Qwen3 runtime measured in Figure~\ref{fig:efficiency}(b,c) retains native
FlashAttention over the paged complete KV cache. Local reads a persistent
ring workspace containing the same selected history within its access range.
For $s=4$ and $w=2{,}048$ (including the current token), it holds at most
2,051 committed positions:
four initial positions in dedicated slots and 2,047 recent positions in a ring.
The current Local candidate KV is supplied separately and remains uncommitted until
selection, preserving the pre-step history for either branch.
Advancing the ring replaces its oldest recent entry; the complete paged cache
retains that entry for later Full access, so total KV storage grows with context.
Queries and keys use their original absolute positions for rotary encoding.
Ring wraparound changes physical storage slots while preserving token positions
and cached key encodings.

\paragraph{Conditional execution and fused commitment.}
Local, the actual FP32 recall head, conditional Full, and commitment run
inside device-side CUDA Graphs. Four captured stages---Local plus the head,
Full, Local commitment, and Full commitment---form two mutually exclusive
branches after scoring. Each step uses one top-level graph launch, which
executes multiple kernels. The device predicate applies the score
threshold and nonfinite fallback, so routing requires no per-token host
score read. A Full retry reads the same preceding committed KV as Local.
If Local is selected, one fused kernel writes its current KV across all
layers to both the paged cache and the appropriate Local slots. If Full
is selected, native Full has already written its current paged KV; the
fused update copies the Full candidate into the Local workspace without
writing the paged cache a second time. Unselected scratch is overwritten
by the next Local attempt. Thus, both history views advance with the same
selected candidate once per step. Updating the selected hidden state and
advancing the position are separate graph operations. The common engine
output path projects and samples once outside this conditional graph;
engine scheduling and CPU sequence bookkeeping remain part of decoding.

\paragraph{Length-compatible graph reuse.}
In the measured vLLM~0.18 Qwen3-1.7B implementation, conditional graph
groups are keyed by $\lceil L/128\rceil$, where $L$ is the effective
attention length including the current token. Each group is captured at
an actual length in its interval. For the pinned FlashAttention backend
and model shape, the KV block count, split plan, and rounded length are
compatible throughout that interval. Replay updates the device-side
effective length and current cache position. Reuse also checks model
parameters, the routing threshold, input and metadata tensor addresses
and layouts, and the attention backend;
using the engine's maximum capacity as a universal capture length would
not preserve the same attention reduction plan.

Crossing an interval boundary transfers the preceding selected hidden
state and routing counters on the GPU. Capture and warm-up save and
restore the current paged slot, ring slot, hidden state,
and counters, so trial executions do not advance the committed history.
This rollback is confined to graph construction; ordinary replay updates
only the current selected state.
Each new request initializes its own history and
head state; reusing workspaces or compatible graphs does not reuse the
previous request's history. In tensor-parallel execution, graph capture
uses the vLLM TP-group context and registers custom-allreduce buffers for
graph replay. The warm-up and timing protocol is given in
Appendix~\ref{app:efficiency}.

When Local and Full have identical access sets, the Transformers
implementation can reuse the Local result for a logical Full action.
The distinct-range cost model assumes $n>s+w$.

\paragraph{Direct-ring Local implementation.}
Local attention uses two Triton kernels that directly read ring KV.
The first processes blocks of 16 logical positions, maps query heads to KV
heads for GQA, and computes block maxima, exponential sums, and weighted
values in FP32. The second combines these partial results with stable
softmax rescaling and casts the output to BF16. Persistent per-layer
scratch buffers support up to 129 blocks, avoiding repeated allocation,
window gathering, and KV-head replication. GPU position tensors determine
ring addresses at replay time.

The standalone Local reference uses the same attention kernels with a
single graph for embedding, the Local backbone, and ring commitment. It
has no recall head or complete paged history after Full prefill. Its
execution costs therefore differ from forcing zero Full calls in ODA.

\paragraph{Hybrid recurrent state.}
For Qwen3.5-2B, Local restricts only the six ordinary attention layers,
at zero-based indices 3, 7, 11, 15, 19, and 23. The remaining 18 Gated
DeltaNet layers retain their native computation. Before Local, the runtime
saves convolution and recurrent states; before a Full retry, it restores
the pre-step state. Only the selected updates are committed. Native
precision uses BF16 backbone, KV, and convolution state, with FP32
recurrent state. These state transactions are necessary for recomputation
from the same history; the Qwen3 timing results do not establish a speedup
for this hybrid implementation.

\subsection{Cost of conditional recomputation}
\label{app:speed-model}

Let $B$ be history-independent backbone cost, $A(n)$ attention cost at
history length $n$, $R$ recall-head cost, and $O$ the shared output cost.
Let $D_F$ and $D_{\mathrm{ODA}}$ denote execution overheads. For a fixed
Local range $\ell=s+w$ and recall rate $\rho$, an additive timing model is
\begin{align}
    T_F(n)&=B+A(n)+O+D_F,\\
    T_{\mathrm{ODA}}(n,\rho)
    &=B+A(\ell)+R+\rho[B+A(n)]+O+D_{\mathrm{ODA}}.
    \label{eq:speed-general}
\end{align}
The Local attempt is paid at every step, including recalled steps, while
the output projection is paid only once. This model assumes no reused
backbone prefix and comparable branch kernels; measured branch times can
replace the shared $B,A$ approximation.

Writing $H=R+D_{\mathrm{ODA}}-D_F$, acceleration requires
\begin{equation}
    \rho<\frac{A(n)-A(\ell)-H}{B+A(n)}.
    \label{eq:speed-budget}
\end{equation}
For $A(n)=an$, $a>0$, and $\rho<1$, this becomes
$n>[\rho B+a\ell+H]/[(1-\rho)a]$.
The ratio approaches $1/\rho$ only when global-attention cost dominates
all other terms. Thus, a 12.5\% recall rate does not imply an $8\times$
finite-context speedup, and a fixed positive recall rate retains $O(n)$
expected attention cost per decoding step.

For an actual continuation, costs must be summed over its history lengths
$n_t$ and recall indicators $a_t$:
\begin{equation}
    T_{\mathrm{ODA}}=\sum_t
    \bigl[B_t+A(\ell_t)+R_t+a_t(B_t+A(n_t))+O_t+D_{\mathrm{ODA},t}\bigr].
    \label{eq:speed-sequence}
\end{equation}
The timing of recall matters through the lengths $n_t$, not only its
frequency. Similarly, total attended positions count both Local reads and
any additional Full reads. Neither total traffic nor throughput can be
inferred from $1-\rho$ alone. The FLOPs measurements in
Appendix~\ref{app:efficiency} use the actual model shapes and call schedules.

\section{Additional Experiments and Evaluation Protocols}
\label{app:experiments}

\subsection{Training and evaluation settings}
\label{app:evaluation-details}
\label{app:scale}

\paragraph{Training runs.}
The recall heads are trained with frozen backbones. Qwen heads use
model-tokenized versions of a content-deduplicated long-context SFT corpus
derived from Step-3.5-Flash-SFT~\citep{step35sft2026}. The training pool
contains 196,608 examples with maximum length 16,384; development and
test splits each contain 8,192 examples. Pool size is not the number of
distinct examples consumed during a training run.
The main-table operating points are separate, model-specific runs. The
Qwen3-1.7B head used for the RULER16K main result is also used for
the five-task Random control and the fixed-reference diagnostics.
Its LongBench main result uses a separate head. These trained heads are
reported as measured operating points rather than selected by an
independent validation rule.
The primary Qwen3-1.7B training configuration and measured resource
consumption are detailed in Appendix~\ref{app:training-cost}.

\paragraph{Generation protocol.}
RULER16K uses 13 tasks with 100 examples each. LongBench v1 is the
ten-task, 1,950-example evaluation under a 40,960-token input budget, rather
than the full LongBench collection. Scores follow each task's official
metric and are macro-averaged within each benchmark. The reported NPU
evaluations use greedy decoding with thinking disabled, Full prefill,
and a 512-token output cap. Each policy advances its own generated
continuation and selected cache. StreamingLLM and ODA's Local branch use four
initial positions and a 2,048-token recent window including the current position. In hybrid
backbones, the restriction applies only to layers that otherwise read
the full history; native local or Gated DeltaNet layers retain their
own computation.

\paragraph{Full-call rates and aggregation.}
Full-call rate is the fraction of routed decoding steps assigned to Full,
excluding the first output produced by prefill. For task $d$, let $F_d$
and $N_d$ be the Full-decision and routed-step counts summed across its
examples, giving $\rho_d=F_d/N_d$. Across $D$ tasks, we distinguish
\[
    \rho_{\mathrm{pooled}}=\frac{\sum_d F_d}{\sum_d N_d},
    \qquad
    \rho_{\mathrm{task}}=\frac{1}{D}\sum_d\rho_d.
\]
The pooled rate weights tasks by routed-step count; the task-averaged rate
weights tasks equally. RULER main-table rates at each length, input
ablations, and training-duration comparisons use the pooled rate.
RULER's Avg averages the five length-specific rates equally. LongBench
and the five-task ODA--Random comparison use the task-averaged rate.
Random means and standard deviations are computed across seed-level
statistics, as detailed in Appendix~\ref{app:budget-controls}.
Individual trajectory rates use that trajectory's Full-decision and
routed-step counts. Runtime experiments use prescribed Full-call rates
over a fixed continuation, such as $126/1008=12.5\%$.
The offline selection budget in fixed-reference diagnostics instead
specifies the fraction of reference positions selected for access;
it is distinct from a free-generation Full-call rate.

\paragraph{Evaluation cohorts and source precision.}
The Qwen3-1.7B RULER16K point uses a GPU evaluation, together with
Full and StreamingLLM from the same evaluation cohort. The remaining
main-table points use NPU evaluations.
RULER benchmark scores and Full-call rates at 4k, 8k, 32k, and 64k are
transcribed from the original result tables at two-decimal precision for
all five models and three methods. The five-length averages combine these
values with the higher-precision 16k records.
Qwen3-1.7B uses a different recall head for its 16k main-table point
than for the other RULER lengths. Its five-length Avg therefore combines
distinct operating points and is descriptive.
The other models each use the same recall head and fixed threshold
across RULER lengths.
In particular, the 32k--64k comparisons use an unchanged head and threshold
within each model.
Qwen3-8B and Qwen3.5-27B share heads across the two benchmarks;
Qwen3-1.7B and Gemma use benchmark-specific heads.
The supplemental Qwen3.5 LongBench ODA task scores and Full-call rates are
transcribed from the original result tables at two-decimal precision.
Gemma's LongBench Full and StreamingLLM
references use batch 2, while the ODA run uses batch 1. Its score
differences are therefore descriptive cross-batch comparisons.

\subsection{Additional token-level generation trajectories}
\label{app:generation-trajectories}

\begin{figure}[htbp]
\centering
\includegraphics[width=\linewidth]{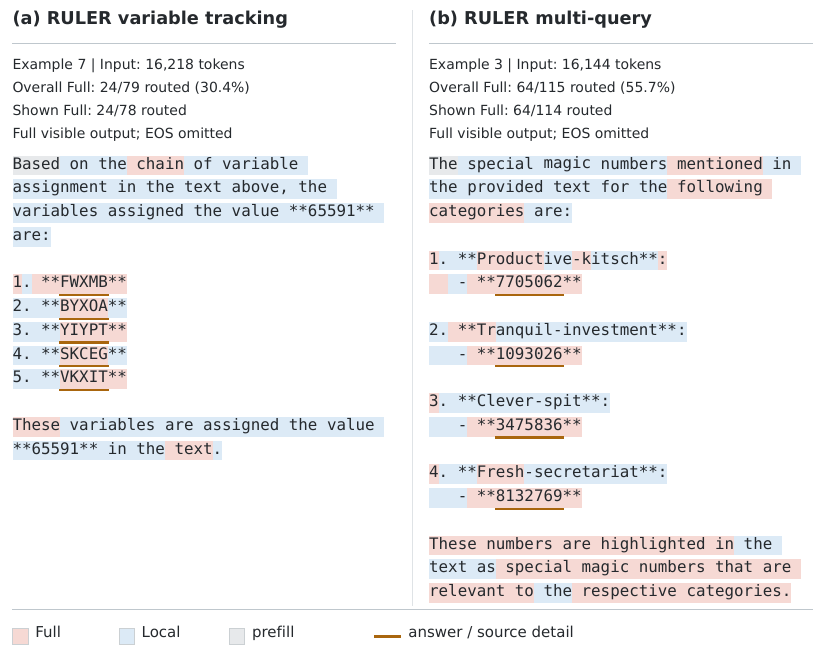}
\caption{\textbf{Additional token-level access allocation on RULER16K.}
\textbf{(a)} Variable tracking, example 7: all 16 tokens overlapping the five
underlined reference variable names use Full, while the complete trajectory
uses Full at 24 of 79 routed positions (30.4\%).
\textbf{(b)} Multi-query retrieval, example 3: all 28 tokens overlapping the
four underlined reference numbers use Full, while the complete trajectory
uses Full at 64 of 115 routed positions (55.7\%).
Both panels show the complete visible output with the termination token
omitted. Local handles many positions in the surrounding explanation
and list syntax.}
\label{fig:appendix-retrieval-trajectories}
\end{figure}

\begin{figure}[p]
\centering
\includegraphics[width=\linewidth]{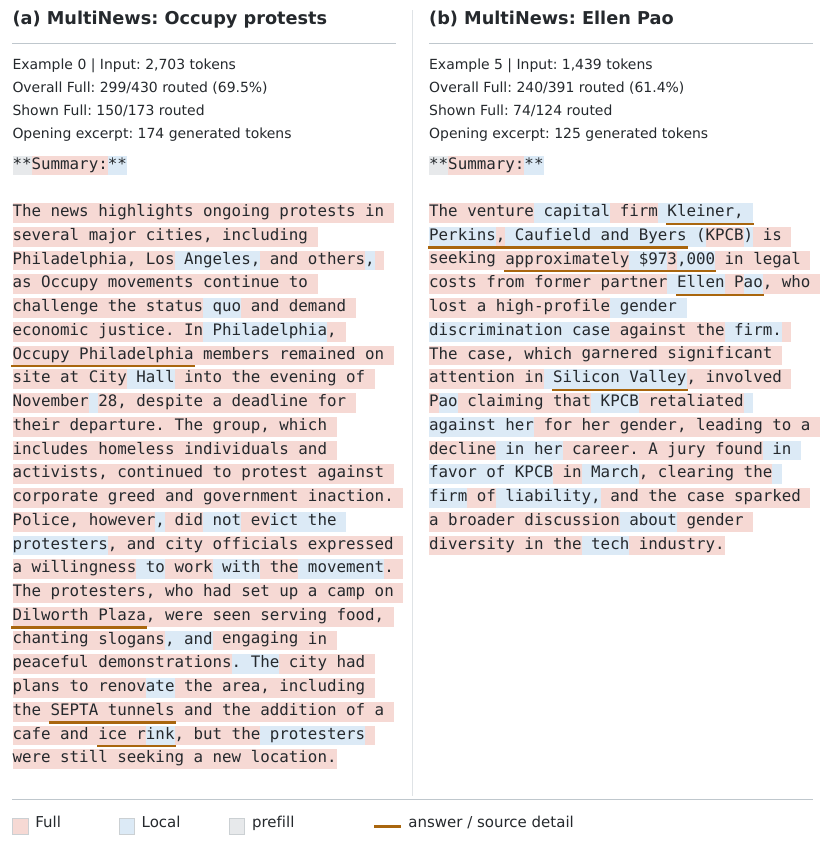}
\caption{\textbf{Variation in token-level access allocation within MultiNews.}
Both panels show contiguous opening paragraphs from qualitative summary replays.
\textbf{(a)} Occupy protests, example 0: the first 174 generated tokens
contain 150 Full calls over 173 routed positions; the complete trajectory
uses 299/430 (69.5\%). Full is selected for 11 of 12 tokens overlapping the
underlined source-linked details.
\textbf{(b)} Ellen Pao case, example 5: the first 125 generated tokens
contain 74 Full calls over 124 routed positions; the complete trajectory
uses 240/391 (61.4\%). Only 4 of 25 tokens overlapping the marked entities
and figures use Full. This example starts with 1,439 input tokens and
reaches a maximum context of 1,830 tokens, so its complete history fits
within the 2,048-token Local window. It illustrates recorded routing when
the complete history is locally available.}
\label{fig:appendix-summary-trajectories}
\end{figure}

Figures~\ref{fig:appendix-retrieval-trajectories}
and~\ref{fig:appendix-summary-trajectories} extend
Figure~\ref{fig:generation-trajectories} with four examples from the same
frozen Qwen3-1.7B head used for the RULER16K main result, at threshold zero.
All examples use greedy decoding, thinking disabled, and a 512-token output cap.
RULER examples come from the recorded RULER16K rerun; QMSum and MultiNews
are qualitative replays with the same frozen head and runtime;
the LongBench main table uses a separately trained head.
Archived uninstrumented replay checks reproduce generated token IDs and
native routing counts for both MultiNews cases.

Pink, blue, and gray denote recorded Full decisions, Local decisions, and
the initial token produced by prefill. Orange underlines mark reference
answers in RULER or manually selected source-linked summary details.
The main RULER panel shows the complete visible answer without its termination
token; the QMSum panel shows the first 85 tokens, ending at a sentence boundary.
Full-rate denominators exclude the prefill token. Complete-trace counts
include the termination token; the panels separately report the counts
for the displayed output. These selected examples describe observed
allocation and its variation across outputs, without establishing that
Full was necessary at each marked position.

\clearpage

\subsection{Per-task generation results}
\label{app:task-results}

The following tables give per-task scores for each completed main-table
operating point, with evaluation settings given
in Appendix~\ref{app:evaluation-details}.
LongBench uses the same ten tasks and order as its main
table. Each table pairs Full and StreamingLLM with ODA for the same model and task
set. The overall ODA Full-call rate appears after Avg in smaller indigo type.
LongBench appendix averages and rates therefore match the main table.
The Qwen3-1.7B and Qwen3-8B LongBench per-task Full-call rates were supplied
from their batch-2 run records to two decimal places.
Task scores and benchmark averages are displayed to two decimal places.
Averages use the full stored precision where available.
For the supplemental Qwen3.5 LongBench ODA results, task scores and
Full-call rates are available only to two decimal places, and the reported
averages are computed from those supplied values.

\begin{table}[htbp]
\centering
\small
\setlength{\tabcolsep}{5pt}
\caption{Qwen3-1.7B RULER16K per-task scores.}
\label{tab:per-task-ruler}
\begin{tabular}{@{}lrrr@{}}
\toprule
Task & Full & StreamingLLM & ODA \\
\midrule
cwe & 12.00 & 25.80 & 14.90 \\
fwe & 84.00 & 70.33 & 84.00 \\
niah\_multikey\_1 & 99.00 & 17.00 & 100.00 \\
niah\_multikey\_2 & 89.00 & 10.00 & 88.00 \\
niah\_multikey\_3 & 90.00 & 2.00 & 82.00 \\
niah\_multiquery & 98.25 & 14.75 & 97.25 \\
niah\_multivalue & 97.00 & 14.75 & 94.50 \\
niah\_single\_1 & 100.00 & 12.00 & 100.00 \\
niah\_single\_2 & 100.00 & 18.00 & 100.00 \\
niah\_single\_3 & 100.00 & 11.00 & 100.00 \\
qa\_hotpotqa & 37.00 & 18.00 & 34.00 \\
qa\_squad & 60.00 & 21.00 & 61.00 \\
vt & 99.00 & 15.40 & 99.60 \\
\midrule
Avg & 81.94 & 19.23 & 81.17\odaavgrate{41.64} \\
\bottomrule
\end{tabular}
\end{table}

\begin{table}[htbp]
\centering
\small
\setlength{\tabcolsep}{5pt}
\caption{Qwen3-8B RULER16K per-task scores.}
\label{tab:per-task-ruler16k-qwen38b}
\begin{tabular}{@{}lrrr@{}}
\toprule
Task & Full & StreamingLLM & ODA \\
\midrule
cwe & 91.40 & 46.90 & 88.90 \\
fwe & 94.00 & 86.00 & 93.00 \\
niah\_multikey\_1 & 99.00 & 17.00 & 99.00 \\
niah\_multikey\_2 & 98.00 & 10.00 & 99.00 \\
niah\_multikey\_3 & 98.00 & 5.00 & 95.00 \\
niah\_multiquery & 100.00 & 14.50 & 99.75 \\
niah\_multivalue & 100.00 & 12.50 & 99.75 \\
niah\_single\_1 & 100.00 & 12.00 & 100.00 \\
niah\_single\_2 & 100.00 & 18.00 & 100.00 \\
niah\_single\_3 & 100.00 & 12.00 & 100.00 \\
qa\_hotpotqa & 53.00 & 40.00 & 50.00 \\
qa\_squad & 79.00 & 41.00 & 79.00 \\
vt & 96.40 & 16.00 & 93.20 \\
\midrule
Avg & 92.98 & 25.45 & 92.05\odaavgrate{54.17} \\
\bottomrule
\end{tabular}
\end{table}

\begin{table}[htbp]
\centering
\small
\setlength{\tabcolsep}{5pt}
\caption{Qwen3.5-2B RULER16K per-task scores.}
\label{tab:per-task-ruler16k-qwen352b}
\begin{tabular}{@{}lrrr@{}}
\toprule
Task & Full & StreamingLLM & ODA \\
\midrule
cwe & 97.00 & 84.40 & 97.60 \\
fwe & 100.00 & 92.00 & 99.33 \\
niah\_multikey\_1 & 100.00 & 17.00 & 100.00 \\
niah\_multikey\_2 & 100.00 & 13.00 & 100.00 \\
niah\_multikey\_3 & 100.00 & 5.00 & 99.00 \\
niah\_multiquery & 100.00 & 14.75 & 100.00 \\
niah\_multivalue & 100.00 & 15.00 & 98.00 \\
niah\_single\_1 & 100.00 & 16.00 & 100.00 \\
niah\_single\_2 & 100.00 & 18.00 & 100.00 \\
niah\_single\_3 & 100.00 & 13.00 & 100.00 \\
qa\_hotpotqa & 51.00 & 24.00 & 51.00 \\
qa\_squad & 78.00 & 31.00 & 78.00 \\
vt & 100.00 & 24.40 & 100.00 \\
\midrule
Avg & 94.31 & 28.27 & 94.07\odaavgrate{44.06} \\
\bottomrule
\end{tabular}
\end{table}

\begin{table}[htbp]
\centering
\small
\setlength{\tabcolsep}{5pt}
\caption{Qwen3.5-27B RULER16K per-task scores.}
\label{tab:per-task-ruler16k-qwen3527b}
\begin{tabular}{@{}lrrr@{}}
\toprule
Task & Full & StreamingLLM & ODA \\
\midrule
cwe & 77.10 & 82.00 & 100.00 \\
fwe & 99.67 & 89.33 & 99.67 \\
niah\_multikey\_1 & 100.00 & 18.00 & 99.00 \\
niah\_multikey\_2 & 100.00 & 16.00 & 100.00 \\
niah\_multikey\_3 & 100.00 & 5.00 & 99.00 \\
niah\_multiquery & 100.00 & 12.50 & 93.25 \\
niah\_multivalue & 100.00 & 12.00 & 87.75 \\
niah\_single\_1 & 100.00 & 16.00 & 100.00 \\
niah\_single\_2 & 100.00 & 16.00 & 100.00 \\
niah\_single\_3 & 100.00 & 13.00 & 100.00 \\
qa\_hotpotqa & 74.00 & 54.00 & 73.00 \\
qa\_squad & 96.00 & 67.00 & 94.00 \\
vt & 100.00 & 6.80 & 99.60 \\
\midrule
Avg & 95.91 & 31.36 & 95.79\odaavgrate{29.47} \\
\bottomrule
\end{tabular}
\end{table}

\begin{table}[htbp]
\centering
\small
\setlength{\tabcolsep}{5pt}
\caption{Gemma-4-12B-it RULER16K per-task scores.}
\label{tab:per-task-ruler16k-gemma412bit}
\begin{tabular}{@{}lrrr@{}}
\toprule
Task & Full & StreamingLLM & ODA \\
\midrule
cwe & 99.80 & 90.50 & 99.80 \\
fwe & 99.67 & 80.33 & 99.67 \\
niah\_multikey\_1 & 100.00 & 18.00 & 100.00 \\
niah\_multikey\_2 & 100.00 & 15.00 & 100.00 \\
niah\_multikey\_3 & 100.00 & 5.00 & 99.00 \\
niah\_multiquery & 100.00 & 15.25 & 99.00 \\
niah\_multivalue & 99.50 & 6.00 & 94.75 \\
niah\_single\_1 & 100.00 & 16.00 & 100.00 \\
niah\_single\_2 & 100.00 & 16.00 & 100.00 \\
niah\_single\_3 & 100.00 & 13.00 & 100.00 \\
qa\_hotpotqa & 70.00 & 46.00 & 70.00 \\
qa\_squad & 91.00 & 51.00 & 91.00 \\
vt & 96.00 & 18.80 & 96.40 \\
\midrule
Avg & 96.61 & 30.07 & 96.12\odaavgrate{67.50} \\
\bottomrule
\end{tabular}
\end{table}

\begin{table}[htbp]
\centering
\small
\setlength{\tabcolsep}{5pt}
\caption{Qwen3-1.7B LongBench v1 per-task scores.}
\label{tab:per-task-longbench}
\begin{tabular}{@{}lrrr@{}}
\toprule
Task & Full & StreamingLLM & ODA \\
\midrule
qasper & 37.34 & 31.69 & 36.84 \\
multifieldqa\_en & 46.78 & 30.42 & 45.67 \\
multifieldqa\_zh & 55.29 & 39.39 & 54.99 \\
2wikimqa & 34.62 & 29.46 & 34.59 \\
hotpotqa & 39.60 & 31.10 & 40.00 \\
musique & 16.06 & 8.36 & 16.28 \\
dureader & 32.42 & 17.75 & 31.61 \\
qmsum & 22.76 & 20.92 & 22.47 \\
vcsum & 15.15 & 13.52 & 14.87 \\
passage\_retrieval\_zh & 95.00 & 48.00 & 94.67 \\
\midrule
Avg & 39.50 & 27.06 & 39.20\odaavgrate{80.12} \\
\bottomrule
\end{tabular}
\end{table}

\begin{table}[htbp]
\centering
\small
\setlength{\tabcolsep}{5pt}
\caption{Qwen3-8B LongBench v1 per-task scores.}
\label{tab:per-task-longbench10-qwen38b}
\begin{tabular}{@{}lrrr@{}}
\toprule
Task & Full & StreamingLLM & ODA \\
\midrule
qasper & 46.08 & 37.59 & 45.96 \\
multifieldqa\_en & 54.13 & 37.30 & 53.83 \\
multifieldqa\_zh & 62.55 & 45.74 & 62.55 \\
2wikimqa & 40.79 & 33.55 & 40.78 \\
hotpotqa & 56.76 & 45.91 & 56.77 \\
musique & 32.08 & 21.67 & 31.89 \\
dureader & 26.78 & 17.68 & 26.62 \\
qmsum & 24.19 & 21.53 & 24.03 \\
vcsum & 13.94 & 14.20 & 14.05 \\
passage\_retrieval\_zh & 97.50 & 49.50 & 97.50 \\
\midrule
Avg & 45.48 & 32.47 & 45.40\odaavgrate{88.35} \\
\bottomrule
\end{tabular}
\end{table}

\begin{table}[htbp]
\centering
\small
\setlength{\tabcolsep}{5pt}
\caption{Qwen3.5-2B LongBench v1 per-task scores.}
\label{tab:per-task-longbench10-qwen352b}
\begin{tabular}{@{}lrrr@{}}
\toprule
Task & Full & StreamingLLM & ODA \\
\midrule
qasper & 39.63 & 32.56 & 39.21 \\
multifieldqa\_en & 52.06 & 34.55 & 51.75 \\
multifieldqa\_zh & 61.61 & 44.04 & 61.72 \\
2wikimqa & 34.63 & 29.98 & 34.63 \\
hotpotqa & 48.15 & 36.21 & 48.14 \\
musique & 31.92 & 17.64 & 32.29 \\
dureader & 27.97 & 16.17 & 28.65 \\
qmsum & 22.07 & 19.95 & 21.87 \\
vcsum & 13.13 & 11.87 & 12.73 \\
passage\_retrieval\_zh & 84.50 & 50.50 & 84.50 \\
\midrule
Avg & 41.57 & 29.35 & 41.55\odaavgrate{82.23} \\
\bottomrule
\end{tabular}
\end{table}

\begin{table}[htbp]
\centering
\small
\setlength{\tabcolsep}{5pt}
\caption{Qwen3.5-27B LongBench v1 per-task scores.}
\label{tab:per-task-longbench10-qwen3527b}
\begin{tabular}{@{}lrrr@{}}
\toprule
Task & Full & StreamingLLM & ODA \\
\midrule
qasper & 47.63 & 41.36 & 47.18 \\
multifieldqa\_en & 55.24 & 40.28 & 55.02 \\
multifieldqa\_zh & 64.73 & 53.37 & 64.89 \\
2wikimqa & 66.79 & 64.01 & 66.70 \\
hotpotqa & 67.58 & 58.73 & 67.65 \\
musique & 55.71 & 47.17 & 55.75 \\
dureader & 24.07 & 16.27 & 23.29 \\
qmsum & 22.23 & 19.72 & 22.13 \\
vcsum & 13.11 & 12.45 & 12.94 \\
passage\_retrieval\_zh & 100.00 & 98.00 & 100.00 \\
\midrule
Avg & 51.71 & 45.14 & 51.55\odaavgrate{53.07} \\
\bottomrule
\end{tabular}
\end{table}

\begin{table}[htbp]
\centering
\small
\setlength{\tabcolsep}{5pt}
\caption{Gemma-4-12B-it LongBench v1 per-task scores.}
\label{tab:per-task-longbench10-gemma412bit}
\begin{tabular}{@{}lrrr@{}}
\toprule
Task & Full & StreamingLLM & ODA \\
\midrule
qasper & 48.06 & 41.81 & 48.23 \\
multifieldqa\_en & 58.14 & 42.83 & 56.63 \\
multifieldqa\_zh & 66.62 & 51.71 & 65.61 \\
2wikimqa & 65.50 & 60.68 & 65.68 \\
hotpotqa & 63.70 & 54.02 & 63.74 \\
musique & 46.59 & 33.47 & 46.01 \\
dureader & 27.18 & 16.68 & 25.87 \\
qmsum & 24.22 & 20.85 & 24.14 \\
vcsum & 14.72 & 13.45 & 14.26 \\
passage\_retrieval\_zh & 100.00 & 98.50 & 99.50 \\
\midrule
Avg & 51.47 & 43.40 & 50.97\odaavgrate{58.65} \\
\bottomrule
\end{tabular}
\end{table}

\clearpage

\subsection{Random and learned global access}
\label{app:budget-controls}

\paragraph{Common evaluation.}
The Qwen3-1.7B head from the RULER16K main result is evaluated on NIAH Single,
NIAH Multikey, variable tracking, frequent-word extraction, and QA SQuAD,
with the same 100 examples per task across policies. This is a separate
NPU evaluation with batch size one, a BF16 backbone, an FP32 head, greedy
decoding, and thinking disabled. Full prefill, the Local range, and
selected-state commitment are shared. Full and StreamingLLM are reevaluated under
this setup, scoring 88.626 and 27.214 on the five-task macro.

\paragraph{Random control.}
ODA uses thresholds $-0.01,-0.005,-0.001,0,0.001,0.005,0.01,0.05$.
For each threshold and task $d$, Random uses the unrounded observed ODA
Full rate $\rho_d^{\mathrm{ODA}}$ as its Bernoulli probability at each
routed step, with seeds 42, 43, and 44. Both policies then follow their
own generated trajectories. This controls the task-specific target
frequency, without guaranteeing identical realized rates. No reference
answer selects Random's individual actions.

Both plotted coordinates average the five tasks equally. Each Random
seed first supplies a task-averaged score and Full-call rate; reported
means and error bars are their means and sample standard deviations over
the three routing seeds. They do not measure prompt or training-seed
uncertainty. At threshold zero, the plotted task-averaged Full-call rates
are 41.33\% for ODA and 40.99\% for Random. The corresponding pooled
Full-call rates are 36.86\% and 36.06\%, respectively. Random's pooled
rate is computed separately within each seed and then averaged across
the three seeds; counts are not pooled across seeds.

\begin{table}[htbp]
\centering
\small
\caption{Learned and random access for eight ODA thresholds. Random entries
are means $\pm$ sample standard deviations across three routing seeds.
Scores and realized Full-call rates are averaged across tasks.}
\label{tab:budget-controls}
\begin{tabular}{@{}rrrrr@{}}
\toprule
Threshold & ODA score & ODA Full (\%) & Random score & Random Full (\%) \\
\midrule
-0.01 & 89.05 & 58.29 & $35.75 \pm 0.86$ & $58.03 \pm 0.33$ \\
-0.005 & 88.95 & 47.09 & $33.56 \pm 0.65$ & $46.61 \pm 0.14$ \\
-0.001 & 89.32 & 42.12 & $32.90 \pm 1.03$ & $41.72 \pm 0.21$ \\
0 & 88.96 & 41.33 & $32.79 \pm 0.95$ & $40.99 \pm 0.09$ \\
0.001 & 89.09 & 40.57 & $32.64 \pm 0.81$ & $40.28 \pm 0.19$ \\
0.005 & 88.99 & 37.82 & $32.34 \pm 0.29$ & $37.47 \pm 0.13$ \\
0.01 & 88.88 & 36.13 & $32.32 \pm 0.19$ & $35.88 \pm 0.17$ \\
0.05 & 85.17 & 24.85 & $30.25 \pm 0.95$ & $24.81 \pm 0.55$ \\
\bottomrule
\end{tabular}

\end{table}

\begin{table}[htbp]
\centering
\small
\caption{Per-task scores at ODA threshold zero. Parentheses report actual
Full-call percentages; Random entries average three routing seeds.}
\label{tab:random-per-task}
\begin{tabular}{@{}lrrrr@{}}
\toprule
Task & Full & StreamingLLM & ODA (Full\%) & Random (Full\%) \\
\midrule
NIAH Single & 100.00 & 12.00 & 100.00 (31.04\%) & 12.00 (30.92\%) \\
NIAH Multikey & 98.00 & 17.00 & 99.00 (38.26\%) & 17.67 (38.32\%) \\
Variable tracking & 98.80 & 15.40 & 99.80 (33.62\%) & 16.60 (33.60\%) \\
Frequent words & 84.33 & 70.67 & 83.00 (42.58\%) & 76.67 (42.20\%) \\
QA SQuAD & 62.00 & 21.00 & 63.00 (61.15\%) & 41.00 (59.89\%) \\
\bottomrule
\end{tabular}

\end{table}

All 170 task runs are retained: 40 ODA, 120 Random, five Full, and five
StreamingLLM runs,
covering 17,000 generations. Eight Random task runs depart from the target
rate by more than two percentage points, all on QA SQuAD; the maximum is
3.71 points. Output-cap hits are 9/4,000 for ODA, 75/12,000 for Random,
1/500 for Full, and 2/500 for StreamingLLM. Thus, the observed score gap includes
the effects of routing on generation length and termination. Runner
elapsed times are not used as optimized-vLLM measurements.

\begin{figure}[htbp]
\centering
\includegraphics[width=\linewidth]{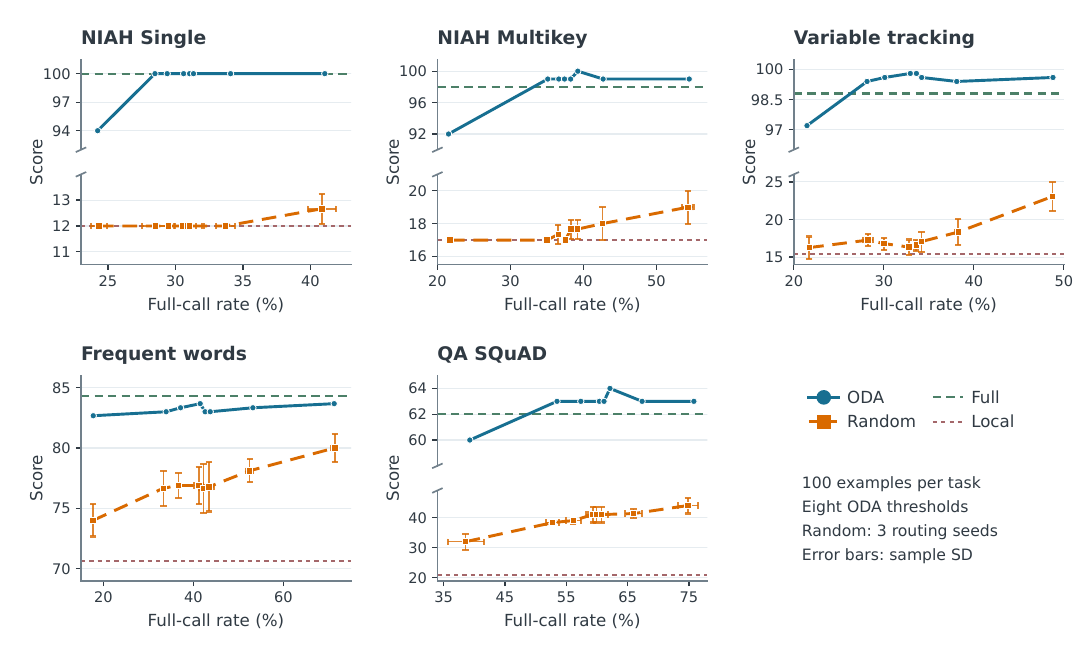}
\caption{\textbf{Learned and random access across individual tasks.}
Random error bars show standard deviations across three routing seeds
on both axes. Panels use task-specific axis ranges; vertical axes are
broken except for frequent-word extraction.}
\label{fig:random-per-task}
\end{figure}

\subsection{Input composition and training behavior}
\label{app:local-state-ablation}

\paragraph{Input subsets.}
Seven heads use subsets of the preceding selected hidden state ($H$),
current token embedding ($E$), and current Local hidden state ($L$).
This ablation series uses a separate training configuration from the
primary head in Table~\ref{tab:main-results}.
All start from the same seed-42 initialization and retain the same
28,325,889-parameter architecture. Missing streams are filled by the first
available input in $H,E,L$ order:
$H\mapsto(H,H,H)$, $E\mapsto(E,E,E)$, $L\mapsto(L,L,L)$,
$HE\mapsto(H,E,H)$, $HL\mapsto(H,H,L)$, $EL\mapsto(E,E,L)$,
and $HEL\mapsto(H,E,L)$. Each head is trained separately and every policy
computes Local before routing. In particular, the $L$-only head can request
Full and is distinct from the fixed StreamingLLM baseline.

All seven input-ablation heads use a frozen BF16 backbone, an FP32 head,
eight-way data parallelism, global batch 64, microbatch one per device,
and eight accumulation steps. Training uses AdamW~\citep{adamw2019},
peak learning rate $3\times10^{-4}$, a 102-step warm-up followed by
cosine decay, gradient clipping at 1.0, and zero head dropout and weight
decay. Their cost penalty is $\lambda=0.001$ and history threshold is
$\tau_{\mathrm{hist}}=0$. Evaluation uses Transformers SDPA, batch size two, greedy decoding,
Full prefill, and at most 512 outputs on the same 13 RULER16K tasks,
with 100 examples per task.
Every trained head uses threshold zero. Since the resulting Full rates differ,
these comparisons measure the learned policies' natural operating points;
they do not isolate an input's contribution at equal access frequency.

Table~\ref{tab:state-ablation} reports these seven input configurations
after 1,024 training updates. Each configuration has one training trajectory
with the shared initialization seed.

The complete $H+E+L$ run continues with the same optimizer, schedule,
random state, and data order. Its Full rate falls from 76.22\% at update
1,024 to 50.11\% at 2,048, then rises to 71.43\% at 3,072, while its
scores remain between 81.33 and 81.57. Longer training does not monotonically
improve this run's quality--access trade-off.

\begin{table}[htbp]
\centering
\small
\caption{Effect of training updates on the complete $H+E+L$ head.}
\label{tab:hel-training-duration}
\begin{tabular}{@{}rrrrr@{}}
\toprule
Updates & RULER16K & Full (\%) & Full calls & Routed steps \\
\midrule
1,024 & 81.57 & 76.22 & 60,848 & 79,829 \\
2,048 & 81.33 & 50.11 & 40,634 & 81,083 \\
3,072 & 81.36 & 71.43 & 59,098 & 82,739 \\
\bottomrule
\end{tabular}

\end{table}

\subsection{Fixed-reference Qwen3 diagnostics}
\label{app:qwen3-reference}

\paragraph{Collection.}
The Qwen3-1.7B head from the RULER16K main result is evaluated with CPU FP32/SDPA.
Sixteen test sequences are chosen by fixed content-hash order, four in each
length bucket with edges 4,096, 6,144, 8,192, 10,240, and 16,385. Each
contributes one contiguous block of 128 supervised reference positions
beyond the Local window. Selection does not use predictions or gains;
the supplied content audit finds none of these sequences in train or dev.
After Full prefill, reference tokens advance ODA's selected history. Both
current candidates start from the same saved state, and only the branch
chosen by the head is committed. The reference token is fixed independently
of that action. Observer-disabled replay matches the actions, selected
top-1 tokens, and final KV digests on all 16 sequences.

\paragraph{Benefit ranking.}
Positive benefit is $g_t>0$. Local-correct and Local-incorrect subsets
are determined after evaluation from the Local top-1 prediction. They are
not available to the policy. Table~\ref{tab:qwen3-reference-discrimination}
compares the head with Local entropy. On the 675 Local-incorrect positions,
the other scalar confidence rankings---negative maximum probability,
negative probability margin, and negative top-two log-probability ratio---
also have AUROCs between 0.486 and 0.491. Directions are fixed to rank
lower-confidence positions first.

\begin{table}[t]
\caption{Benefit-sign discrimination on Qwen3-1.7B fixed-reference positions.
Positive benefit is $g_t>0$; subsets are defined by Local's top-1 correctness.
Intervals are 95\% paired sequence-bootstrap intervals.}
\label{tab:qwen3-reference-discrimination}
\centering
\small
\begin{tabularx}{\linewidth}{@{}Xrrll@{}}
\toprule
Subset & Positions & $g_t>0$ & Router AUROC [95\% CI] & Entropy AUROC [95\% CI] \\
\midrule
All & 2,048 & 52.0\% & 0.556 [0.502, 0.610] & 0.534 [0.455, 0.618] \\
Local correct & 1,373 & 52.4\% & 0.503 [0.445, 0.564] & 0.593 [0.483, 0.689] \\
Local incorrect & 675 & 51.1\% & 0.643 [0.577, 0.707] & 0.486 [0.442, 0.526] \\
\bottomrule
\end{tabularx}
\end{table}

For offline budget $b$, each method selects its top
$K=\lfloor b|\mathcal S|\rfloor$ positions within population $\mathcal S$.
Net gain is $|\mathcal S|^{-1}\sum_{t\in\mathcal A}g_t$, including negative
gains. Random's expectation is $(K/|\mathcal S|)$ times the population mean
gain; Oracle ranks by realized gain and uses information unavailable online.
At 40\% of all 2,048 positions, $K=819$: ODA, Random, and Oracle capture
0.149978, 0.053385, and 0.227453 nats per position. The ODA--Random
interval is $[0.045747,0.160134]$. Intervals use 2,000 paired document-bootstrap
resamples with seed 20260910, repeating selection within each resample.

Table~\ref{tab:qwen3-reference-gain} separates positive gains from negative
costs. Across all positions, the head exceeds entropy in 40\%-budget net
gain, with difference interval $[0.0315,0.1388]$, although their AUROC
and AP differences remain unresolved. Within Local-incorrect positions,
the AUROC difference interval is $[0.069,0.230]$ and the net-gain difference
interval is $[0.1516,0.5010]$. Within Local-correct positions, both rankings
have negative net gain. These are complete-head comparisons on fixed
histories, rather than a confidence-controlled attribution to input $L$.

\begin{table}[htbp]
\caption{Full--Local gain decomposition for Qwen3-1.7B.
Each subset is ranked independently, selecting 40\% of its positions.
Positive gain and negative cost are sums in nats; mean net gain is
normalized by the subset's total number of reference positions.}
\label{tab:qwen3-reference-gain}
\centering
\small
\begin{tabular}{@{}llrrrl@{}}
\toprule
Subset & Ranking & Calls & Positive gain & Negative cost & Mean net gain [95\% CI] \\
\midrule
All & Router & 819 & 378.486 & 71.331 & 0.1500 [0.0679, 0.2497] \\
All & Entropy & 819 & 283.170 & 149.730 & 0.0652 [0.0114, 0.1265] \\
Local correct & Router & 549 & 10.253 & 23.822 & $-0.0099$ [$-0.0186$, $-0.0026$] \\
Local correct & Entropy & 549 & 19.336 & 40.529 & $-0.0154$ [$-0.0284$, $-0.0046$] \\
Local incorrect & Router & 270 & 378.419 & 50.544 & 0.4857 [0.2503, 0.7118] \\
Local incorrect & Entropy & 270 & 163.380 & 52.520 & 0.1642 [0.0663, 0.2647] \\
\bottomrule
\end{tabular}
\end{table}

\paragraph{Benefit is distinct from correctness.}
Of 2,048 positions, 1,342 are top-1 correct under both branches, 31 only
under Local, 60 only under Full, and 615 under neither. Among the 1,342
both-correct positions, Full increases the reference probability at 718
and decreases it at 437; the remaining 187 gains are zero. Among the 675
Local-incorrect positions, 345 have positive gain but only 60 become top-1
correct under Full. Local can therefore have a better immediate prediction
even though Full has lower aggregate NLL.

Table~\ref{tab:gain-sensitivity} reports sign counts at several magnitude
cutoffs. The cutoffs are descriptive, not validated numerical-noise bounds.
Changing the positive label from $g_t>0$ to $g_t>0.001$ leaves the
Local-incorrect AUROC and AP unchanged because that subset contains no
gains in $(0,0.001]$. The CPU precision and fixed-reference protocol are
separate from the GPU free-generation evaluations.

\begin{table}[htbp]
\caption{Reference-position counts at different NLL-gain thresholds for
Qwen3-1.7B. Each row partitions the same 2,048 positions;
$\epsilon$ is measured in nats.}
\label{tab:gain-sensitivity}
\centering
\begin{tabular}{@{}rrrr@{}}
\toprule
$\epsilon$ & $g_t<-\epsilon$ & $|g_t|\leq\epsilon$ & $g_t>\epsilon$ \\
\midrule
0 & 797 & 187 & 1,064 \\
0.0001 & 649 & 632 & 767 \\
0.001 & 594 & 788 & 666 \\
0.01 & 503 & 1,012 & 533 \\
0.1 & 313 & 1,419 & 316 \\
\bottomrule
\end{tabular}
\end{table}

\subsection{Runtime protocol and complete results}
\label{app:efficiency}

\paragraph{Environment and workload.}
The matrix uses vLLM 0.18.0, A100 GPUs, batch size one, and pipeline
parallelism one. Backbone and KV are BF16; the recall head is FP32.
The 4K--128K groups use one GPU (TP1), 256K uses two (TP2), and 512K uses
four (TP4). All five methods within each length run on the same physical
GPU group; speedups are normalized to that group's native Full baseline.
Thus, the longer-context points also change the hardware allocation.
YaRN~\citep{yarn2024} uses factors 5, 9, and 17 with maximum sequence
capacities of 139,264, 278,528, and 540,672 for the 4K--128K, 256K, and
512K groups, respectively. Each group's methods share the same RoPE
configuration. These extensions support timing and do not establish
model quality at those lengths.

Each length uses one complete benchmark prompt, without repetition or
concatenation: 4,139 tokens from ZeroSCROLLS Qasper~\citep{zeroscrolls2023},
15,918 from LongBench v2~\citep{longbenchv2_2025},
31,400 from LongBench v1 NarrativeQA~\citep{longbench2024},
63,852 from ZeroSCROLLS NarrativeQA,
128,852 from LooGLE long-dependency summarization~\citep{loogle2024},
and 261,696 and 516,788 from RULER multi-query and multi-key retrieval,
respectively. Within each group, all methods share the prompt and a fixed
1,009-token continuation produced by Full. The recall
head runs after every ODA Local attempt, but a device-side schedule selects
Full for the first 2, 4, or 8 steps of every 16-step period. Over 1,008
routed steps, this gives 126, 252, or 504 Full calls, exactly 12.5\%,
25\%, or 50\%. Nonfinite scores trigger Full and fail the count check.
Native Full uses vLLM's \texttt{FULL\_DECODE\_ONLY} CUDA Graph with
compilation mode zero. Standalone Local uses its own whole-step graph,
without a recall head or complete paged history after Full prefill;
it is distinct from zero-call ODA.

\paragraph{Timing.}
Throughput is $1,008/(t_{1009}-t_1)$, where timestamps record when each
output becomes available from the serial vLLM engine loop. It includes
all layers, the head, conditional Full, state commitment, output
projection, controlled output selection, sampling, and the engine loop.
It excludes loading, tokenization, prefill and its first output,
detokenization, and HTTP transport. Numerical observers are off.
Each configuration runs one complete warm-up request to populate all
relevant graphs and then three measured requests. Reported TPS is their
median; speedup is the ratio of method and Full medians. Capture counters
must remain unchanged during measured Local and ODA requests.

\paragraph{Major-operation FLOPs.}
Using the notation of Appendix~\ref{app:speed-model}, with costs now counted
in FLOPs, the ledger is
\begin{align}
    F_{\mathrm{Full}}&=\sum_t[B+A(n_t)+O],\\
    F_{\mathrm{Local}}&=\sum_t[B+A(\ell_t)+O],\\
    F_{\mathrm{ODA}}&=\sum_t[(1+a_t)B+A(\ell_t)+a_tA(n_t)+R+O].
\end{align}
The count includes MLP and QKV/output projections, both attention branches,
the recall head, and one vocabulary projection. It uses actual model shapes,
history lengths, and schedules. Attention FLOPs scale with query heads;
KV bytes scale with KV heads. This is an analytical count of major
operations, not measured hardware FLOPs or a latency prediction, and
excludes tensor-parallel communication and execution-management costs.
At 516,788 input tokens and 126 Full calls, the computation ratio is
6.375 and the measured warm throughput ratio is 2.650. The full matrix
retains slowdowns at short contexts and larger recall budgets.

\begin{table}[p]
\centering
\footnotesize
\setlength{\tabcolsep}{4pt}
\caption{Runtime and major FLOPs for all 35 Qwen3-1.7B vLLM configurations.
TP denotes the number of tensor-parallel GPUs. TPS is tokens/s, reported
as the median of three requests after warm-up.
Speedup and FLOPs savings use the corresponding native Full baseline.}
\label{tab:decode-throughput-full}
\label{tab:decode-throughput}
\begin{tabular}{@{}rrlrrrr@{}}
\toprule
Input tokens & TP & Method & Full (\%) & TPS & Speedup & FLOPs saved (\%) \\
\midrule
4,139 & 1 & Full & 100.0 & 205.46 & $1.000\times$ & 0.00 \\
4,139 & 1 & Local & 0.0 & 213.90 & $1.041\times$ & 13.19 \\
4,139 & 1 & ODA & 12.5 & 178.52 & $0.869\times$ & 1.17 \\
4,139 & 1 & ODA & 25.0 & 164.90 & $0.803\times$ & -9.60 \\
4,139 & 1 & ODA & 50.0 & 143.00 & $0.696\times$ & -31.15 \\
\addlinespace[3pt]
15,918 & 1 & Full & 100.0 & 177.30 & $1.000\times$ & 0.00 \\
15,918 & 1 & Local & 0.0 & 214.27 & $1.209\times$ & 45.73 \\
15,918 & 1 & ODA & 12.5 & 175.80 & $0.992\times$ & 33.53 \\
15,918 & 1 & ODA & 25.0 & 159.46 & $0.899\times$ & 22.11 \\
15,918 & 1 & ODA & 50.0 & 136.67 & $0.771\times$ & -0.73 \\
\addlinespace[3pt]
31,400 & 1 & Full & 100.0 & 148.31 & $1.000\times$ & 0.00 \\
31,400 & 1 & Local & 0.0 & 214.21 & $1.444\times$ & 63.64 \\
31,400 & 1 & ODA & 12.5 & 171.73 & $1.158\times$ & 51.34 \\
31,400 & 1 & ODA & 25.0 & 153.73 & $1.037\times$ & 39.57 \\
31,400 & 1 & ODA & 50.0 & 126.68 & $0.854\times$ & 16.01 \\
\addlinespace[3pt]
63,852 & 1 & Full & 100.0 & 111.55 & $1.000\times$ & 0.00 \\
63,852 & 1 & Local & 0.0 & 215.00 & $1.927\times$ & 78.51 \\
63,852 & 1 & ODA & 12.5 & 163.84 & $1.469\times$ & 66.13 \\
63,852 & 1 & ODA & 25.0 & 140.90 & $1.263\times$ & 54.06 \\
63,852 & 1 & ODA & 50.0 & 111.37 & $0.998\times$ & 29.91 \\
\addlinespace[3pt]
128,852 & 1 & Full & 100.0 & 75.27 & $1.000\times$ & 0.00 \\
128,852 & 1 & Local & 0.0 & 214.45 & $2.849\times$ & 88.19 \\
128,852 & 1 & ODA & 12.5 & 150.69 & $2.002\times$ & 75.75 \\
128,852 & 1 & ODA & 25.0 & 123.74 & $1.644\times$ & 63.49 \\
128,852 & 1 & ODA & 50.0 & 89.67 & $1.191\times$ & 38.96 \\
\addlinespace[3pt]
261,696 & 2 & Full & 100.0 & 76.44 & $1.000\times$ & 0.00 \\
261,696 & 2 & Local & 0.0 & 274.81 & $3.595\times$ & 93.85 \\
261,696 & 2 & ODA & 12.5 & 174.00 & $2.276\times$ & 81.38 \\
261,696 & 2 & ODA & 25.0 & 142.31 & $1.862\times$ & 69.00 \\
261,696 & 2 & ODA & 50.0 & 98.87 & $1.293\times$ & 44.25 \\
\addlinespace[3pt]
516,788 & 4 & Full & 100.0 & 77.64 & $1.000\times$ & 0.00 \\
516,788 & 4 & Local & 0.0 & 342.63 & $4.413\times$ & 96.80 \\
516,788 & 4 & ODA & 12.5 & 205.71 & $2.650\times$ & 84.31 \\
516,788 & 4 & ODA & 25.0 & 158.02 & $2.035\times$ & 71.88 \\
516,788 & 4 & ODA & 50.0 & 107.29 & $1.382\times$ & 47.01 \\
\bottomrule
\end{tabular}

\end{table}

\paragraph{Numerical checks and scope.}
Implementation checks first compare the Local kernels with an independent
FP32 materialized-window attention calculation on the same Q/K/V inputs.
They then compare graph execution against eager execution with the same
Local kernels, fixed continuation, and routing schedule. Logit checks
require cosine similarity at least 0.999, KL at most 0.005, and top-1
agreement at least 0.99; repeated request resets must match exactly.
These checks cover attention computation and selected-state execution,
rather than equivalence to the Always-Full output distribution.
The measurements isolate execution cost at controlled call schedules.
Joint task quality and speed for the same learned policy, prefill-inclusive
latency, peak memory, and multi-request batch scaling have not been measured.

\clearpage
\subsection{Primary Qwen3-1.7B training and resources}
\label{app:training-cost}

\paragraph{Data and sampling.}
The primary run uses Step-3.5-Flash-SFT~\citep{step35sft2026}
restricted to its \texttt{compiled/qwen3/general} subset.
We retain complete records with 4,096--16,384 tokens, without truncation.
Candidates are randomly ordered within four length bins,
$[4096,6144)$, $[6144,8192)$, $[8192,10240)$, and $[10240,16385)$,
using seed 20260721. After global exact-content SHA-256 deduplication and
exclusion of the fixed development and test sets (8,192 records each),
each bin contributes 49,152 records. The resulting 196,608-record dataset
contains 1,663,312,927 non-padding input tokens. A dataset audit identifies
886,844,336 eligible supervision positions under the four-sink,
2,048-token-window configuration.

\paragraph{Optimization.}
The BF16 backbone remains frozen, and only the FP32 recall head's
28,325,889 parameters are optimized. Training uses eight A100-SXM4-80GB
GPUs with DDP, microbatch size two per GPU, and four accumulation steps,
giving a global batch of 64. AdamW~\citep{adamw2019} uses zero weight decay,
gradient-norm clipping at 1.0, and a learning rate that warms up for 102
updates to $3\times10^{-4}$ and then remains constant through update
3,072. This run completes exactly one epoch; its configured cosine phase
begins at the stopping boundary. The training seed is 20260827. Training uses
FlexAttention with PyTorch 2.7.1, CUDA 12.6, and Triton 3.3.1.
These settings describe the original primary run; the input-ablation
configuration is reported separately in Appendix~\ref{app:local-state-ablation}.

\paragraph{Measured resources.}
Summing the recorded step times over all 3,072 updates gives 13.67 hours,
or 109.34 GPU-hours across eight GPUs. The interval from the first startup
log to the final checkpoint save is 13.75 hours, corresponding to an
eight-GPU allocation of approximately 110 GPU-hours. These totals cover
the complete one-epoch run, including frozen-backbone feature and paired-label
computation; the primary RULER16K result uses the head after 1,024 updates.
Data preparation, preliminary
experiments, other training runs, and downstream evaluation are excluded.

\end{document}